\documentclass[11pt]{article}

\usepackage[preprint]{acl}

\usepackage{times}
\usepackage{latexsym}
\usepackage{amssymb}
\usepackage{amsmath}
\usepackage{multirow}
\usepackage{booktabs}
\usepackage{subcaption}
\usepackage{makecell}
\usepackage{pifont}
\usepackage[table]{xcolor}
\usepackage{enumitem}
\usepackage{algorithm}
\usepackage{algpseudocode}

\usepackage[T1]{fontenc}

\usepackage[utf8]{inputenc}

\usepackage{microtype}

\usepackage{inconsolata}

\usepackage{graphicx}

\definecolor{darkred}{RGB}{204, 0, 0}
\definecolor{forestgreen}{RGB}{34, 139, 34}

\newcommand{\declinecell}[2]{\makecell[c]{#1 (\textcolor{darkred}{$\downarrow$#2})}}
\newcommand{\efficiencycell}[2]{\makecell[c]{#1 \\ \textcolor{forestgreen}{$\downarrow$#2}}}
\newcommand{\cmark}{\ding{51}}
\newcommand{\xmark}{\ding{55}}

\newcommand{\textbfit}[1]{\textbf{\textit{#1}}}

\newcommand{\eoi}{\texttt{[EOI]}}
\newcommand{\eos}{\texttt{[EOS]}}
\newcommand{\ourname}{{MV-VisRAG}}

\title{When Visually Rich Documents Meet Multi-Vector Retrieval for Region-level Retrieval-Augmented Generation}

\author{
 \textbf{Xu Yuan},
 \textbf{Liangbo Ning},
 \textbf{Yujuan Ding},
 \textbf{Qing Li\thanks{Corresponding Authors.}},
 \textbf{Wenqi Fan\footnotemark[1]},
 \\ \\
 The Hong Kong Polytechnic University, HK SAR\\
 \small{xuyuan127@gmail.com, biglemon1123@gmail.com, dingyujuan385@gmail.com,}\\
 \small{qing-prof.li@polyu.edu.hk, wenqifan03@gmail.com}
}

\makeatletter
\g@addto@macro\normalsize{%
  \setlength{\abovedisplayskip}{4pt}%
  \setlength{\belowdisplayskip}{4pt}%
  \setlength{\abovedisplayshortskip}{2pt}%
  \setlength{\belowdisplayshortskip}{2pt}%
}
\makeatother

\begin{document}
\maketitle
\begin{abstract}
Visually rich documents (VRDs) convey knowledge through textual, visual, and layout information that text-centric Retrieval-Augmented Generation (Text-RAG) struggles to fully exploit.
Recent Vision-RAG approaches enable retrieval and generation directly from document images, retaining this multimodal information.
However, earlier Vision-RAG methods commonly adopt coarse retrieval representations and full-page visual contexts, which can obscure local evidence and introduce substantial redundancy and computational overhead.
To address these challenges, we propose \ourname{}, a region-level Vision-RAG framework that uses semantically complete regions for both document representation and visual augmentation.
Specifically, \ourname{} first constructs these regions through a Content-Adaptive Regionalization module. 
It then encodes them into local region vectors and a page-level global vector using a VLM-based retriever trained through multimodal reconstruction pretraining and retrieval fine-tuning.
With this compact multi-vector representation, \ourname{} performs late-interaction retrieval and estimates region contributions from the resulting matching scores.
These contributions guide the selection of relevant regions, which are then rearranged into a compact visual context for generation.
Experiments across five benchmarks demonstrate retrieval and generation gains over corresponding baselines, while region-level augmentation substantially reduces visual tokens and computational cost. Our code will be made publicly available at \url{https://github.com/xandery-geek/MV-VisRAG}.
\end{abstract}

\section{Introduction}
\begin{figure}
    \centering
    \includegraphics[width=\linewidth]{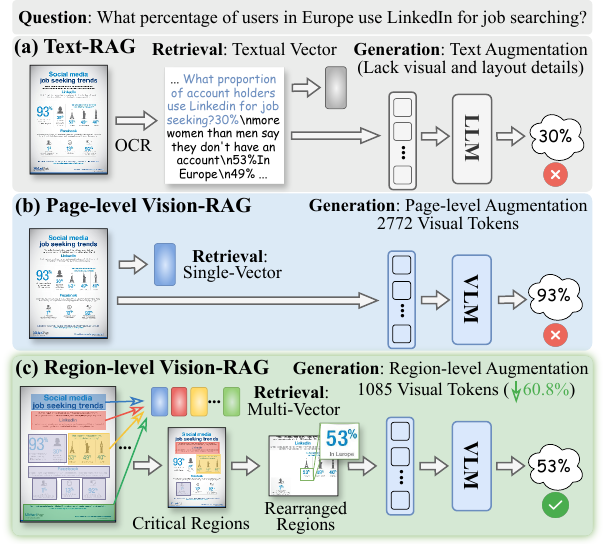}
    \vskip -0.1in
    \caption{Comparison of different RAG paradigms on VRDs.
    \textbf{(a) Text-RAG} relies on parsed text for both retrieval and generation, often introducing noise and neglecting visual and layout information.
    \textbf{(b) Page-level Vision-RAG} operates directly on full-page document images but suffers from coarse-grained retrieval representations and visual augmentation.
    \textbf{(c) Region-level Vision-RAG (Ours)} uses semantically complete regions for multi-vector retrieval and selects high-contribution regions to form a compact visual context for generation.
    }
    \label{fig-motivation}
    \vskip -0.2in
\end{figure}

Retrieval-Augmented Generation (RAG) has become a crucial technique for enhancing Large Language Models (LLMs) by incorporating external knowledge, thereby mitigating hallucinations and the limitations of parameterized knowledge~\cite{lewis2020retrieval,fan2024survey}.
However, most existing RAG frameworks~\cite{ram2023context,asai2024self,wang2025speculative} are predominantly text-centric (referred to as Text-RAG) and struggle to handle Visually Rich Documents (VRDs) that intertwine text and figures, such as textbooks and manuals, where substantial real-world knowledge resides.
As shown in Figure~\ref{fig-motivation}(a), for VRD-based data, Text-RAG typically relies on OCR-based parsing pipelines to acquire textual content.
The parsing process can introduce recognition errors and text--layout misalignment, while text-only representations omit visual information conveyed by figures and charts~\cite{guo2025towards}.

To better leverage VRDs, Vision-RAG uses Vision-Language Models (VLMs)~\cite{liu2023visual,yuan2024instruction,bai2025qwen25vl} to retrieve and answer questions directly from document images, retaining visual and layout information beyond parsed text.
Representative methods such as VisRAG~\cite{yu2025visrag} and VDocRAG~\cite{tanaka2025vdocrag} use page-level retrieval representations and visual contexts, as illustrated in Figure~\ref{fig-motivation}(b). 
Recent work explores finer-grained processing through patch-level multi-vector retrieval~\cite{faysse2025colpali} and region-level visual augmentation~\cite{li2026regionrag}.
However, effectively using fine-grained visual evidence for retrieval and generation still presents two challenges:
\textbf{(1) Compact Fine-grained Retrieval:} Encoding an entire page into a single vector can obscure local evidence, while retaining numerous patch vectors increases storage and retrieval costs. An effective representation should preserve fine-grained matching with a compact document index.
\textbf{(2) Coherent Visual Augmentation:} Full-page inputs contain substantial query-irrelevant content and consume many visual tokens. Selecting smaller regions can reduce this input, but may separate evidence from the surrounding text and layout needed to interpret it. Effective visual augmentation thus requires selecting relevant content while preserving coherent context for answer generation.

To address these challenges, we propose \textbf{\ourname{}}, a region-level Vision-RAG framework that uses semantically complete regions as shared units for indexing, retrieval, and generation.
As illustrated in Figure~\ref{fig-motivation}(c), \ourname{} first constructs query-independent regions from textual and visual layout cues through Content-Adaptive Regionalization (CAR).
To encode these regions for retrieval, we pretrain a VLM-based encoder to reconstruct regional text and page-level visual features, and then fine-tune it for query--document matching.
The resulting local region vectors and page-level global vector form a compact document representation for fine-grained late-interaction retrieval.
Since each local vector corresponds to a document region, the matching scores also indicate which regions contribute to document relevance.
These contributions guide the selection of regions from retrieved documents, which are then rearranged into a compact visual context for generation.
This process preserves the selected regions' content and relative spatial order while reducing query-irrelevant input.

Experiments across five benchmarks show that \ourname{} outperforms single-vector and efficient multi-vector retrievers while offering the most compact multi-vector representations and competitive online retrieval latency.
Our page- and region-level generation variants also outperform their respective Vision-RAG baselines.
Compared with full-page augmentation, the proposed region-level augmentation reduces visual token usage by 43--62\% and lowers generation latency.

Our contributions are summarized as follows:
\begin{itemize}[parsep=0pt, topsep=2pt, itemsep=2pt, leftmargin=12pt]
    \item We present \ourname{}, a region-level Vision-RAG framework that unifies indexing, retrieval, and generation around semantically complete regions constructed by a CAR module.

    \item We develop a region-based multi-vector retriever with multimodal reconstruction pretraining and retrieval fine-tuning, and a selective region-augmented generation strategy that uses region contributions to identify critical visual evidence.

    \item Extensive experiments demonstrate the effectiveness and efficiency of \ourname{} in both retrieval and generation tasks.
\end{itemize}

\begin{figure*}
    \centering
    \includegraphics[width=\linewidth]{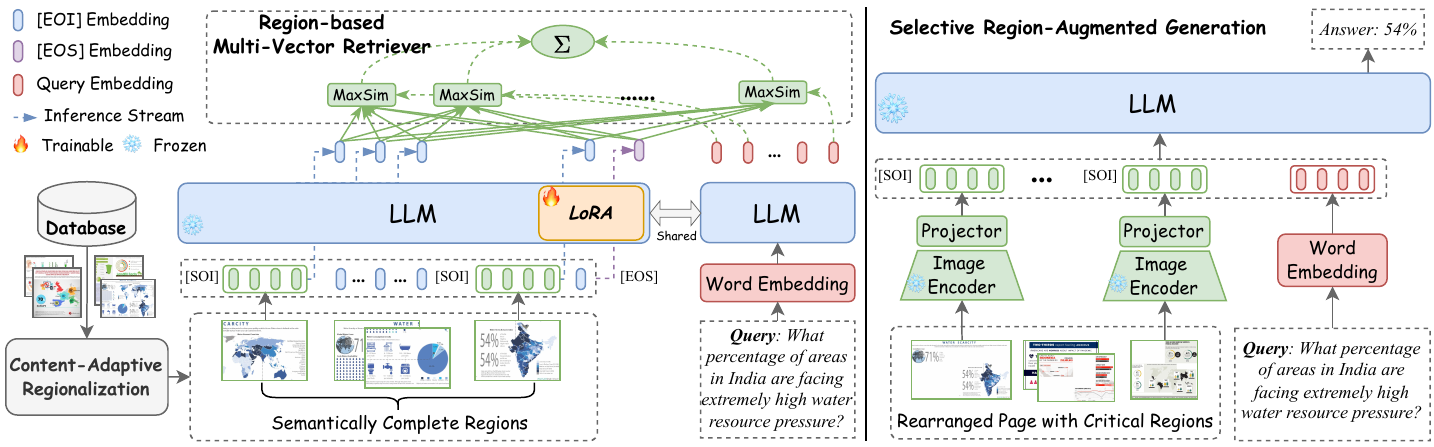}
    \vskip -0.1in
    \caption{Overview of our \ourname{}. \textbf{Left:} \ourname{} performs multi-vector retrieval to identify document images related to the query. \textbf{Right:} \ourname{} leverages selected critical regions to generate the answer.
    }
    \label{fig-framework}
    \vskip -0.2in
\end{figure*}

\section{Related Work}
\noindent \textbf{2.1 Retrieval-augmented Generation.}
RAG has expanded from text domain~\cite{fan2024survey,wei2025instructrag} to heterogeneous knowledge sources, including images~\cite{lin2023fine,yuan2026mkg,yuan2026ai,jiang2026qadragon,jiang2026superglasses} and videos~\cite{luo2026video,liu2025cadencerag}, code~\cite{gao2024preference,dutta2024rar}, and graphs~\cite{wang2025knowledge,wang2026mixture,ding2026fashionkg}.
However, RAG for visually rich documents (VRDs), where knowledge is conveyed jointly by text, layout, tables, and figures, remains underexplored. A central concern is how to represent documents for retrieval and organize relevant evidence for generation.

\noindent \textbf{2.2 Visual Document Retrieval.}
OCR-based pipelines apply textual retrievers~\cite{karpukhin2020dense,wang2024improving,wu2026lineage} to parsed document text, with limited access to visual and layout information.
VLM-based retrievers directly encode document images into page-level single-vector representations~\cite{ma2024unifying,tanaka2025vdocrag}, or patch-level multi-vector representations~\cite{faysse2025colpali}.
To reduce the storage and retrieval overhead of patch-level representations, recent methods compress existing embeddings through pruning~\cite{yan2025docpruner,kankanampati2026voronoi}, merging~\cite{ma2025towards,he2025dual,yuan2026magic}, or their combination~\cite{yan2026sculpting}.
Unlike these post-hoc approaches, \ourname{} constructs semantically complete regions before encoding and directly learns a compact set of region vectors.

\noindent \textbf{2.3 Document Visual Question Answering.}
Document VQA answers questions over documents containing textual, tabular, and visual elements~\cite{mathew2021docvqa,mathew2022infographicvqa,van2023document}.
While conventional methods assume the target document is given~\cite{ye2023ureader}, VisRAG~\cite{yu2025visrag} and VDocRAG~\cite{tanaka2025vdocrag} extend this setting by retrieving relevant document images from a large-scale corpus and using them as full-page visual contexts for generation.
A recent approach, RegionRAG~\cite{li2026regionrag}, uses patch-level embeddings for retrieval and groups salient patches into regions for generation.
\ourname{} instead encodes each semantically complete region into a single vector, reducing index size while preserving coherent regional content as visual evidence for generation.
\section{Methodology}
This section introduces the task definition and an overview of the proposed \ourname{}, followed by a detailed description of each component.

\subsection{Overview}
\noindent \textbf{Task Formulation.}
Given a knowledge-intensive question $Q$ and a set of $N$ visually rich documents $\boldsymbol{D}=\{D_1, ..., D_N\}$, vision-based RAG~\cite{yu2025visrag} requires the model to retrieve related documents as external knowledge and then generate responses.
In general, the task can be decomposed into two key components: 
(1) Document Retriever $\mathcal{R}: (Q, \boldsymbol{D}) \to \boldsymbol{D}_K$, which takes $Q$ and $\boldsymbol{D}$ as inputs and picks top-$K$ candidate documents $\boldsymbol{D}_K \subset \boldsymbol{D}$; 
(2) Answer Generator $\mathcal{G}: (Q, \boldsymbol{D}_K) \to A$, which utilizes both its inner knowledge and outer context $\boldsymbol{D}_K$ to generate answer $A$ for the query $Q$.

\noindent \textbf{The proposed \ourname{}.}
As shown in Figure~\ref{fig-framework}, \ourname{} uses semantically complete regions as shared units for indexing, retrieval, and generation.
First, a Content-Adaptive Regionalization (CAR) module converts the input document into query-independent regions using textual and visual layout cues.
Then, a Region-based Multi-Vector Retriever encodes each document offline into region-level local vectors and a page-level global vector, while representing queries as token-level vectors.
The retriever ranks documents through late interaction and estimates region contributions from the resulting matching scores.
These contributions guide the selection of regions from retrieved documents, which are rearranged into a compact visual context for answer generation.

\subsection{Content-Adaptive Regionalization}
\begin{figure}
    \centering
    \includegraphics[width=\linewidth]{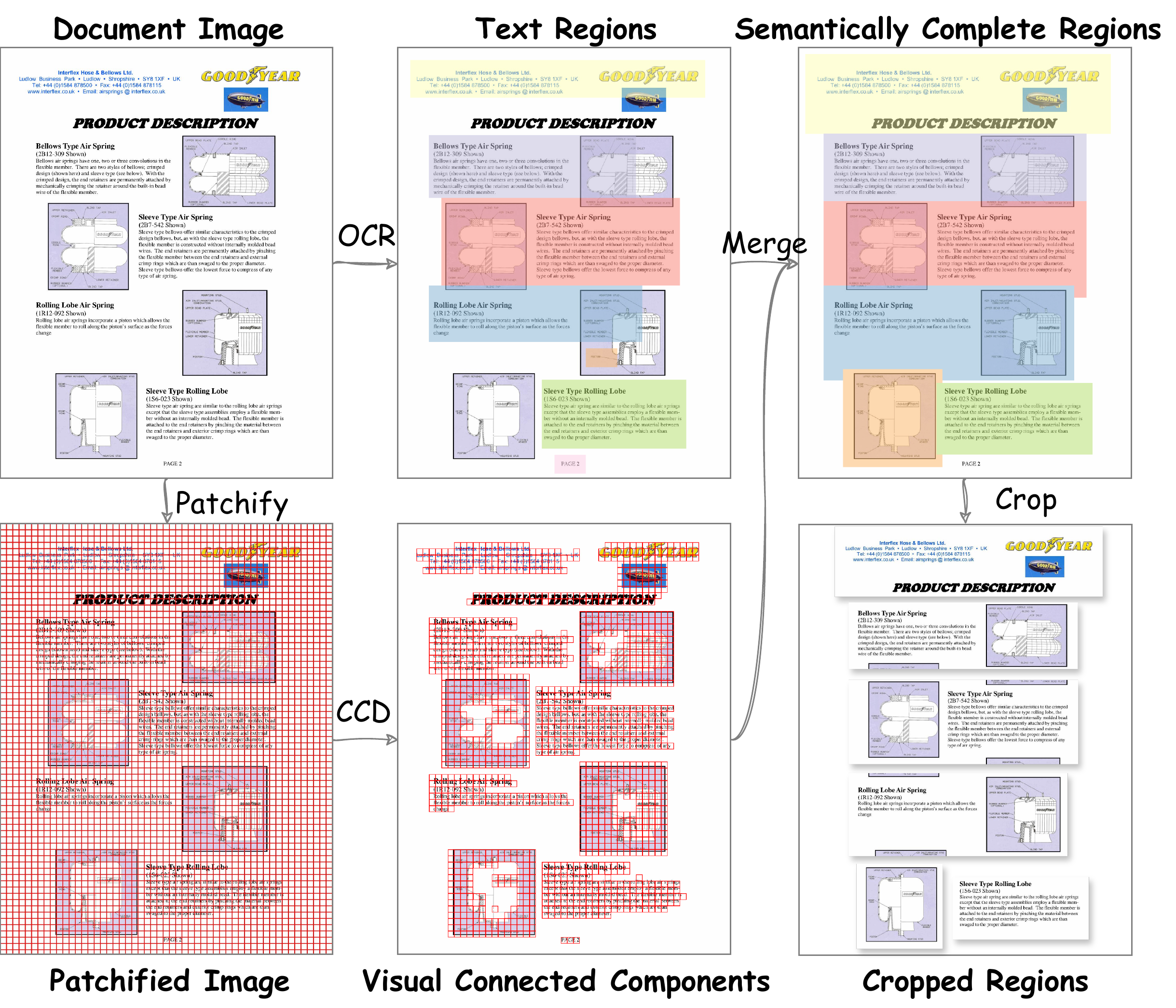}
    \vskip -0.1in
    \caption{Process of Content-Adaptive Regionalization.}
    \label{fig-splitter}
    \vskip -0.2in
\end{figure}

To support both fine-grained retrieval and visual augmentation, document regions should preserve coherent, semantically complete content.
However, patch-level partitioning~\cite{faysse2025colpali} and fixed grid-based splitting~\cite{ye2023ureader, hu2025mplug} may fragment semantically related content across multiple crops. We therefore introduce Content-Adaptive Regionalization (CAR), which integrates text-aware Optical Character Recognition (OCR) with vision-based Connected Component Detection (CCD) to adapt region partitioning to document content, as illustrated in Figure~\ref{fig-splitter}.

\noindent \textbf{Text-aware OCR.}
Textual layout provides structural cues for grouping document content.
Given a document image $D \in \mathbb{R}^{H\times W}$, we extract part-level OCR bounding boxes $\boldsymbol{B} = \{B_1, \ldots, B_L\}$ and iteratively merge spatially close boxes into text regions.
Formally, two boxes $B_i$ and $B_j$ are merged when $d(B_i, B_j) < \tau_{dis}$, where $d(\cdot)$ is the minimum Euclidean distance between their corners in normalized coordinates and $\tau_{dis}$ is the distance threshold.
This process yields a collection of text regions $\boldsymbol{R}_{txt} = \{R_1, \ldots, R_T\}$.

\noindent \textbf{Vision-based CCD.}
Purely OCR-based grouping may overlook non-textual content such as figures and diagrams.
To capture this complementary information, we also recognize visual elements by analyzing pixel-level similarity between neighboring image patches.
Specifically, we divide the document image into a uniform patch grid and model it as a 4-connected graph whose nodes represent patches and edges link neighbors.
Then, a Union--Find strategy is applied to group adjacent patches with similar visual characteristics, yielding a set of visual connected components $\boldsymbol{C}_{vis} = \{C_1, \ldots, C_V\}$. The detailed algorithmic implementation is provided in Appendix~\ref{app-car}.

\noindent \textbf{Finalization with Region Merging.}
We treat text regions $\boldsymbol{R}_{txt}$ as the structural backbone and incorporate overlapping visual components from $\boldsymbol{C}_{vis}$ to bring related textual and visual content together.
Each component $C_v$ is represented by its bounding box and merged with the text region having the highest overlap ratio, provided that this ratio exceeds $\tau_{merge}$.
To control region granularity, regions taller than $\tau_{height}$ are split into horizontal bands, while regions smaller than $\tau_{area}$ are discarded.
The resulting region set $\boldsymbol{R} = \{R_1, \ldots, R_K\}$ is then used for region-based document encoding and subsequent region-level visual augmentation.

\begin{figure*}
    \centering
    \includegraphics[width=0.95\linewidth]{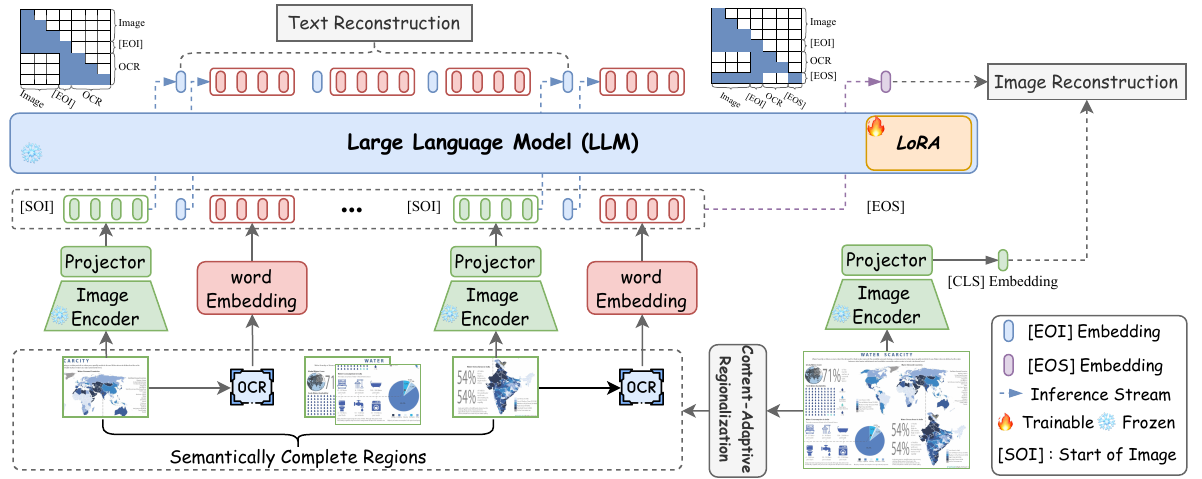}
    \vskip -0.05in
    \caption{The proposed pretraining tasks that learn \eoi{} and \eos{} embeddings via text and image reconstruction.}
    \label{fig-pre-train}
    \vskip -0.15in
\end{figure*}

\subsection{Region-based Multi-Vector Retriever}
We design a region-based multi-vector retriever built on a VLM backbone. 
Although VLMs exhibit strong generative capabilities due to next-token prediction training~\cite{vaswani2017attention}, they often suffer from limited representation power, leading to suboptimal performance on retrieval benchmarks~\cite{lee2025nvembed}. 
To address this issue, we introduce a two-stage training strategy to enhance the representation capabilities of the retriever.

\noindent \textbf{Multi-vector Multimodal Reconstruction Pretraining.}
In the pretraining stage, the retriever learns two complementary representations of document $D$: region-level local embeddings that capture fine-grained textual semantics and a page-level global embedding that preserves the holistic visual context of the entire document.

We first introduce a \textbfit{text reconstruction} task to learn local representations.
As shown in Figure~\ref{fig-pre-train}, we construct an image--text pair $(v_k, t_k)$ for each region $R_k \in \boldsymbol{R}$, where $v_k$ is the region image and $t_k$ is its OCR-extracted text.
We insert an \eoi{} token between $v_k$ and $t_k$ and use its final-layer hidden state, the \eoi{} embedding, as the local representation.
To encourage this embedding to capture textual semantics from $v_k$, we apply a customized attention mask $M$ that allows text tokens in $t_k$ to attend to the corresponding \eoi{} token and preceding text tokens, while blocking direct access to image tokens in $v_k$.
The \eoi{} embedding thus serves as an information bottleneck for autoregressively reconstructing $t_k$ from $v_k$.
Given the token sequence $\{y_1^{k}, \ldots, y_{N^{k}}^{k}\}$ of $t_k$, the local reconstruction loss is defined as:
\begin{equation}
    \mathcal{L}_{local} = - \frac{1}{K} \sum_{k=1}^{K} (\frac{1}{N^{k}} \sum_{i=1}^{N^{k}} \log p(y_i^{k} | \text{[EOI]}^{k}, y_{<i}^{k})).
\end{equation}

To complement regional text semantics with page-level visual information, we append an \eos{} token to the input sequence. 
Its final-layer hidden state, denoted by $h_{eos}$, serves as the global \eos{} embedding. Under the attention mask $M$, this token aggregates visual information from region images without attending to OCR text.
We introduce an \textbfit{image reconstruction} task that trains the \eos{} embedding to capture the visual representation of the holistic document image.
Specifically, we adopt a latent-space reconstruction objective and minimize the $L_2$ distance to the projected class-level embedding of document $D$, i.e.,
\begin{equation}
    \mathcal{L}_{global} = \|h_\text{eos} - \mathcal{\phi}(\mathcal{E}(D)_\text{cls}) \|_{2},
\end{equation}
where $\mathcal{E}$ and $\mathcal{\phi}$ are the frozen visual encoder and projector from the same VLM backbone used by the retriever.
The total pretraining objective is $\mathcal{L}_{pretrain} = \mathcal{L}_{local} + \alpha \mathcal{L}_{global}$, with $\alpha=1.0$.

\noindent \textbf{Multi-Vector Retrieval Fine-tuning.}
In the fine-tuning stage, we continue to refine local and global representations through multi-vector interaction with textual queries, thereby allowing effective retrieval. 
As illustrated in Figure~\ref{fig-framework}, the representation of query $Q$ is derived from the final-layer hidden states, while the document $D$ is represented by the introduced \eoi{} and \eos{} embeddings. 
The late-interaction similarity~\cite{khattab2020colbert} is then computed by summing over all query vectors, taking the maximum dot product for each query vector $q_i$ across all document vectors $d_j$, i.e.,
\begin{equation}
\label{equ-late-interaction}
    s(Q, D) = \sum_{i=1}^{N_q} s_i, \ s_i= \max_{0 \le j \le N_d} \langle q_i , d_j \rangle,
\end{equation}
where $d_0$ and $\{d_1, \ldots, d_{N_d}\}$ denote the global (\eos{}) and local (\eoi{}) representations; $\langle \cdot \rangle$ represents the dot product.
Here, we have $N_d = K$, since each region $R_k$ corresponds to an \eoi{} embedding.
Based on the late-interaction operation, we employ contrastive learning~\cite{chen2020simple,he2020momentum} to encourage positive query-document pairs to have similar features, in contrast to negative pairs in a batch. Let $B$ denote the batch size and $\tau$ be a temperature parameter, the fine-tuning objective is defined as:
\begin{equation}
    \mathcal{L}_{ft} = - \frac{1}{B} \sum_{i=1}^{B} \log \frac{\exp(s(Q_i, D_i)/\tau)}{\Sigma_{j=1}^{B} \exp(s(Q_i, D_j)/ \tau)}
\end{equation}

After fine-tuning, the retriever is used to identify the top-$K_d$ documents relevant to query $Q$.

\begin{table*}[htbp]
\caption{Comparison with textual and single-vector visual retrievers using nDCG@5 ($\uparrow$) and Recall@5 ($\uparrow$).}
\vskip -0.1in
\label{tab-retrieval}
\setlength{\tabcolsep}{3pt}
\resizebox{\textwidth}{!}{
\begin{tabular}{lcc|cccccccccc}
\toprule
\multirow{2}{*}{\textbf{Model}} 
& \multirow{2}{*}{\textbf{Init}} 
& \multirow{2}{*}{\textbf{Scale}} 
& \multicolumn{2}{c}{\textbf{InfoVQA}} 
& \multicolumn{2}{c}{\textbf{DUDE}} 
& \multicolumn{2}{c}{\textbf{DocVQA}} 
& \multicolumn{2}{c}{\textbf{TATDQA}}
& \multicolumn{2}{c}{\textbf{ViDoRe-P}} \\
                                          &             &      & nDCG@5 & R@5 & nDCG@5 & R@5   & nDCG@5 & R@5   & nDCG@5  & R@5 & nDCG@5  & R@5 \\
\midrule
BM25~\cite{robertson2009probabilistic}    & --          & 0    & 33.54 & 40.17 & 42.56 & 48.44 & 56.31 & 62.27 & 54.67 & 66.51 & 82.79 & 89.75 \\
Contriever~\cite{izacard2022unsupervised} & BERT        & 0.1B & 41.86 & 52.19 & 41.39 & 49.29 & 60.60 & 71.40 & 35.37 & 45.04 & 80.20 & 89.00 \\
E5~\cite{wang2022text}                    & BERT        & 0.1B & 48.47 & 58.30 & 45.98 & 54.64 & 62.24 & 70.22 & 33.18 & 42.15 & 83.29 & 91.75 \\
GTE~\cite{li2023towards}                  & BERT        & 0.1B & 50.59 & 60.97 & 43.91 & 53.53 & 60.18 & 69.20 & 29.12 & 36.68 & 80.80 & 88.25 \\
E5-Mistral~\cite{wang2024improving}       & Mistral     & 7.1B & 55.40 & 66.22 & 45.84 & 54.18 & 63.28 & 71.57 & 36.11 & 45.52 & 86.95 & 94.75 \\
NV-Embed-v2~\cite{lee2025nvembed}         & Mistral     & 7.9B & 59.10 & 69.27 & 54.68 & 62.95 & 75.78 & 82.91 & 45.03 & 57.37 & 90.16 & 95.50 \\
\midrule
CLIP~\cite{radford2021learning}           & Scratch     & 0.4B & 41.26 & 51.53 & 24.57 & 29.69 & 28.81 & 38.41 & 10.87 & 14.97 & 50.57 & 60.50 \\
DSE~\cite{ma2024unifying}                 & Phi3V       & 4.2B & 67.12 & 76.72 & 55.51 & 63.96 & 75.82 & 83.59 & 44.43 & 56.22 & 90.02 & 96.00 \\
UniME-V2~\cite{gu2026unime}               & Qwen2-VL    & 7B   & 60.12 & 71.76 & 48.67 & 58.11 & 66.54 & 76.82 & 40.93 & 53.58 & 81.17 & 89.00 \\
VLM2Vec-V2~\cite{meng2026vlmvecv}         & Qwen2-VL    & 2B   & 67.81 & 78.91 & 55.03 & 63.19 & 75.05 & 84.26 & 46.15 & 58.03 & 86.86 & 94.00 \\
VisRAG~\cite{yu2025visrag}                & MiniCPM-V   & 3.4B & 71.80 & 83.11 & 56.35 & 65.94 & 78.00 & 87.48 & 48.89 & 61.27 & 90.05 & 96.00 \\
VDocRAG\dag{}~\cite{tanaka2025vdocrag}    & Phi3V       & 4.2B & 72.93 & 83.07 & 58.52 & 67.44 & 74.79 & 83.76 & 46.15 & 58.99 & 88.04 & 95.00 \\ 
VDocRAG~\cite{tanaka2025vdocrag}          & Qwen2.5-VL  & 3B   & 77.85 & 87.26 & 65.09 & 74.42 & 82.85 & 90.36 & 53.73 & 66.75 & 90.46 & 96.05 \\
\midrule
\ourname{}\dag{} (Ours)                  & Phi3V       & 4.2B  & 73.61 & 83.87 & 60.50 & 69.87 & 78.96 & 88.83 & 54.21 & 66.92 & 90.52 & 96.05 \\
\ourname{} (Ours)                        & Qwen2.5-VL  &   3B  & \textbf{79.36} & \textbf{88.65} & \textbf{70.55} & \textbf{79.76} & \textbf{86.20} & \textbf{93.05} & \textbf{61.37} & \textbf{73.18} & \textbf{92.48} & \textbf{96.75} \\
\bottomrule
\end{tabular}
}
\vskip 0.03in
\footnotesize{\dag{} denotes Phi3V-based variants; unmarked \ourname{} denotes the standard Qwen2.5-VL-based model.}
\end{table*}

\begin{table*}[htbp]
\centering
\caption{Comparison with efficient multi-vector retrievers in nDCG@5 ($\uparrow$) and representation size ($\downarrow$), measured as average embedding number $\times$ dimension. Here, ColQwen is reported as a full-size reference model. Bold marks the best result among efficient multi-vector approaches, each using Qwen2.5-VL-3B as the backbone.}
\vskip -0.1in
\label{tab-retrieval-multi}
\setlength{\tabcolsep}{3pt}
\resizebox{\textwidth}{!}{
\begin{tabular}{l|cccccccccc}
\toprule
\multirow{2}{*}{\textbf{Model}}
& \multicolumn{2}{c}{\textbf{InfoVQA}} 
& \multicolumn{2}{c}{\textbf{DUDE}} 
& \multicolumn{2}{c}{\textbf{DocVQA}} 
& \multicolumn{2}{c}{\textbf{TATDQA}}
& \multicolumn{2}{c}{\textbf{ViDoRe-P}} \\
                                          & nDCG@5 & Size & nDCG@5 & Size    & nDCG@5 & Size   & nDCG@5  & Size & nDCG@5  & Size \\
\midrule
\rowcolor{gray!15}
ColQwen~\cite{faysse2025colpali}          & 80.47 & 723.4$\times$128 & 72.97 & 752.4$\times$128 & 87.88 & 750.9$\times$128 & 63.28 & 752.9$\times$128 & 94.08 & 753.1$\times$128 \\
\midrule
ColQwen-Pooling~\cite{faysse2025colpali}  & 76.88 & 188.9$\times$128 & 67.03 & 188.5$\times$128 & 85.15 & 188.2$\times$128 & 58.34 & 187.8$\times$128 & 92.10 & 188.3$\times$128 \\
DocPruner~\cite{yan2025docpruner}         & 70.90 & 189.3$\times$128 & 66.76 & 196.4$\times$128 & 82.90 & 196.0$\times$128 & 57.65 & 196.5$\times$128 & 89.53 & 196.6$\times$128 \\
Light-ColPali~\cite{ma2025towards}        & 68.67 & 189.8$\times$128 & 40.64 & 189.4$\times$128 & 85.61 & 189.1$\times$128 & 57.73 & 188.7$\times$128 & 92.19 & 189.3$\times$128 \\
RegionRAG~\cite{li2026regionrag}          & 69.69 & 712.4$\times$128 & 56.35 & 741.4$\times$128 & 78.36 & 739.9$\times$128 & 60.73 & 741.9$\times$128 & 82.37 & 742.1$\times$128 \\
PRUNE-THEN-MERGE~\cite{yan2026sculpting}  & 73.67 & 183.3$\times$128 & 69.28 & 190.4$\times$128 & 85.91 & 190.0$\times$128 & 60.97 & 190.5$\times$128 & 90.65 & 190.6$\times$128 \\
\midrule
\ourname{} (Ours)                        
& \textbf{79.36} & \textbf{8.6$\times$2048}  
& \textbf{70.55} & \textbf{7.0$\times$2048}  
& \textbf{86.20} & \textbf{10.8$\times$2048} 
& \textbf{61.37} & \textbf{10.9$\times$2048} 
& \textbf{92.48} & \textbf{8.5$\times$2048}  \\     
\bottomrule
\end{tabular}
}
\vskip -0.15in
\end{table*}

\subsection{Selective Region-Augmented Generation}
\label{sec-selective-rag}
During generation, we further refine each retrieved document by selecting its critical regions that make a remarkable contribution to the late-interaction score, enabling effective and efficient generation.

\noindent \textbf{Region Contribution.}
We estimate region importance from the query--document matches established during late interaction.
For each query vector $q_i$, the index of winning document embedding is
\begin{equation}
    j_i^{*} = \arg \max_{0 \le j \le N_d} \langle q_i, d_j \rangle.
\end{equation}
Using the matching scores $s_i$ defined in Equation~\ref{equ-late-interaction}, we compute the contribution $c_j$ for each document embedding $d_j (j=0,1,\ldots,N_d)$ by accumulating the non-negative scores of the query vectors it wins:
\begin{equation}
    c_j = \sum_{i=1}^{N_q} \boldsymbol{1}\{j_i^{*}=j\}\max(s_i,0),
\end{equation}
where $\boldsymbol{1}\{\cdot\}$ is the indicator function.
The global embedding receives query matches that contribute to page-level relevance, summarized by $c_0$.
To incorporate this signal into region selection, we distribute a weighted share of $c_0$ among local regions according to their similarity to the global embedding.
Finally, the region contribution of region $R_j (j=1,\ldots,N_d)$ is defined as
\begin{equation}
    \tilde{c}_j = c_j +
    \beta \frac{\max(\langle d_j,d_0\rangle,0)}
    {\sum_{\ell=1}^{N_d}\max(\langle d_{\ell},d_0\rangle,0)+\epsilon}c_0,
\end{equation}
where $\beta=0.1$ controls the influence of the global contribution and $\epsilon=10^{-8}$ ensures stability.

\noindent \textbf{Region Selection.}
For each retrieved document, we rank regions by their contributions $\tilde{c}_j$ and use Top-K selection by default, retaining up to $K_r=5$ regions.
We also consider an adaptive strategy based on the Pareto principle~\cite{correa2015pareto}, which assumes that a small fraction of regions accounts for most of the total contribution.
This strategy sorts regions by contribution in descending order and selects the smallest set of top-ranked regions whose cumulative contribution accounts for at least a fraction $\rho$ of the total.

The regions selected by either strategy are then organized into a compact visual context and fed into the VLM for answer generation, as illustrated in the right part of Figure~\ref{fig-framework}.
Specifically, we compose the selected regions from each document into a single image, preserving their relative horizontal and vertical order while removing redundant whitespace, as shown in Figure~\ref{fig-cases}.
Details of region rearrangement are given in Appendix~\ref{app-region-rearrangement}.
\section{Experiments}
\subsection{Setups}

\noindent \textbf{Pretraining and Fine-tuning Datasets.}
Our proposed retriever is trained in two stages. In the pretraining stage, we leverage DocStruct4M~\cite{hu2024mplug}, from which we collect over 300k samples and convert each document image into a sequence of sub-images paired with their corresponding OCR text. The fine-tuning stage is based on OpenDocVQA~\cite{tanaka2025vdocrag}, which contains approximately 41k query–document pairs.

\noindent \textbf{Evaluation Datasets.}
We evaluate Visual Document Retrieval and Document VQA on representative benchmarks covering diverse document types: \textit{InfoVQA}~\cite{mathew2022infographicvqa}, \textit{DUDE}~\cite{van2023document}, \textit{DocVQA}~\cite{mathew2021docvqa}, \textit{TATDQA}~\cite{zhu2022towards}, and the practical subset of \textit{ViDoRe}~\cite{faysse2025colpali}, denoted as \textit{ViDoRe-P}, since its academic tasks largely overlap with the first four benchmarks.
TATDQA and ViDoRe-P are out-of-domain test sets, while the others are used for in-domain evaluation.

\noindent \textbf{Evaluation Metrics.}
We adopt \textit{nDCG@5} and \textit{Recall@5} to evaluate document retrieval performance, following \citet{tanaka2025vdocrag}. For the Document VQA task, we report two standard metrics: \textit{ANLS}~\cite{biten2019scene} and \textit{F1} score.

More details are provided in Appendix~\ref{app-experimental-details}.

\begin{table*}[t]
\caption{Retrieval-augmented generation performance in ANLS and F1. \textbf{Bold} indicates the best within each group.}
\vskip -0.1in
\label{tab-generation}
\resizebox{\textwidth}{!}{
\begin{tabular}{lcc|cccccccccc}
\toprule
\multirow{2}{*}{\textbf{Model}} 
& \multirow{2}{*}{\textbf{Generator}} 
& \multirow{2}{*}{\textbf{Docs}} 
& \multicolumn{2}{c}{\textbf{InfoVQA}} 
& \multicolumn{2}{c}{\textbf{DUDE}} 
& \multicolumn{2}{c}{\textbf{DocVQA}} 
& \multicolumn{2}{c}{\textbf{TATDQA}} 
& \multicolumn{2}{c}{\textbf{ViDoRe-P}} \\
                       &                &                 & ANLS   & F1       & ANLS    & F1      & ANLS    & F1      & ANLS    & F1      & ANLS    & F1    \\
\midrule
Text-RAG           & Qwen2.5      & OCR text        & 39.50 & 21.84 & 52.44 & 44.50 & 64.75 & 56.52 & 44.23 & 30.32 & 49.56 & 53.80 \\
VisRAG             & Qwen2.5-VL   & Page-level      & 63.56 & 52.53 & 45.33 & 34.05 & 82.20 & 78.50 & 43.24 & 27.55 & 53.95 & 57.14 \\
VDocRAG            & Qwen2.5-VL   & Page-level      & 67.64 & 56.73 & 56.50 & 47.97 & 83.65 & 79.70 & 43.37 & 27.34 & 53.51 & 56.47 \\
\ourname{} (Ours)  & Qwen2.5-VL   & Page-level      & \textbf{69.07} & \textbf{58.92} & \textbf{63.18} & \textbf{55.93} & \textbf{84.49} & \textbf{81.26} & \textbf{46.10} & \textbf{31.28} & \textbf{54.01} & \textbf{57.72} \\
\midrule
RegionRAG         & Qwen2.5-VL   & Region-level    & 63.88 & 51.70 & 50.23 & 40.79 & 80.72 & 77.11 & 44.43 & 29.83 & 52.85 & 54.76 \\
\ourname{} (Ours) & Qwen2.5-VL   & Region-level    & \textbf{69.34} & \textbf{59.04} & \textbf{59.66} & \textbf{52.15} & \textbf{82.74} & \textbf{79.73} & \textbf{46.04} & \textbf{31.42} & \textbf{53.87} & \textbf{56.03} \\
\bottomrule
\end{tabular}
}
\vskip -0.1in
\end{table*}

\begin{table}[htbp]
\caption{Visual token counts produced by Qwen2.5-VL under different vision augmentation strategies.}
\vskip -0.1in
\label{tab-generation-efficiency}
\setlength{\tabcolsep}{3pt}
\resizebox{\linewidth}{!}{
\begin{tabular}{c|ccccc}
\toprule
\textbf{Strategy} & \textbf{InfoVQA} & \textbf{DUDE} & \textbf{DocVQA} & \textbf{TATDQA} & \textbf{ViDoRe-P} \\
\midrule
Page-level    & 4292        &  5071          &  5067       &  4844        &   4711     \\
RegionRAG     & 3525        &  3908          &  3880       &  3906        &   3915     \\
\midrule
\ourname{}  
& \efficiencycell{2436}{43.24\%} 
& \efficiencycell{2533}{50.04\%} 
& \efficiencycell{1968}{61.16\%} 
& \efficiencycell{1839}{62.03\%}  
& \efficiencycell{2003}{57.48\%}    \\
\bottomrule
\end{tabular}
}
\vskip -0.15in
\end{table}

\subsection{Retrieval Results}
\label{exp-retrieval}
Table~\ref{tab-retrieval} compares \ourname{} with textual and single-vector visual retrievers.
\ourname{} outperforms textual retrievers, including larger embedding models, demonstrating the value of visual evidence for VRD retrieval.
Compared to VDocRAG, the strongest single-vector visual baseline in the table, \ourname{} improves nDCG@5 and Recall@5 by an average of 3.47 and 3.46 points with Phi3V, and 4.00 and 3.31 points with Qwen2.5-VL, respectively, across the five benchmarks.
By representing each page with multiple region vectors, \ourname{} allows different query tokens to match their relevant local evidence, providing finer-grained query--document alignment.

Table~\ref{tab-retrieval-multi} extends this comparison to efficient multi-vector retrievers, using full-size ColQwen as a reference.
ColQwen achieves the highest nDCG@5 but uses the largest document representations, incurring high storage and retrieval costs.
With the smallest representation size, \ourname{} approaches ColQwen's accuracy and outperforms the other efficient multi-vector methods, providing a favorable balance between retrieval accuracy and storage efficiency.
Further analysis in Appendix~\ref{app-storage_runtime_costs} shows \ourname{} also achieves lower encoding cost and competitive retrieval latency.

In addition, we evaluate \ourname{} on the more recent ViDoRe V2 benchmark in Appendix~\ref{app-vidore_v2}, where it achieves the highest average nDCG@5 and Recall@5.

\begin{figure*}
    \centering
    \includegraphics[width=0.95\linewidth]{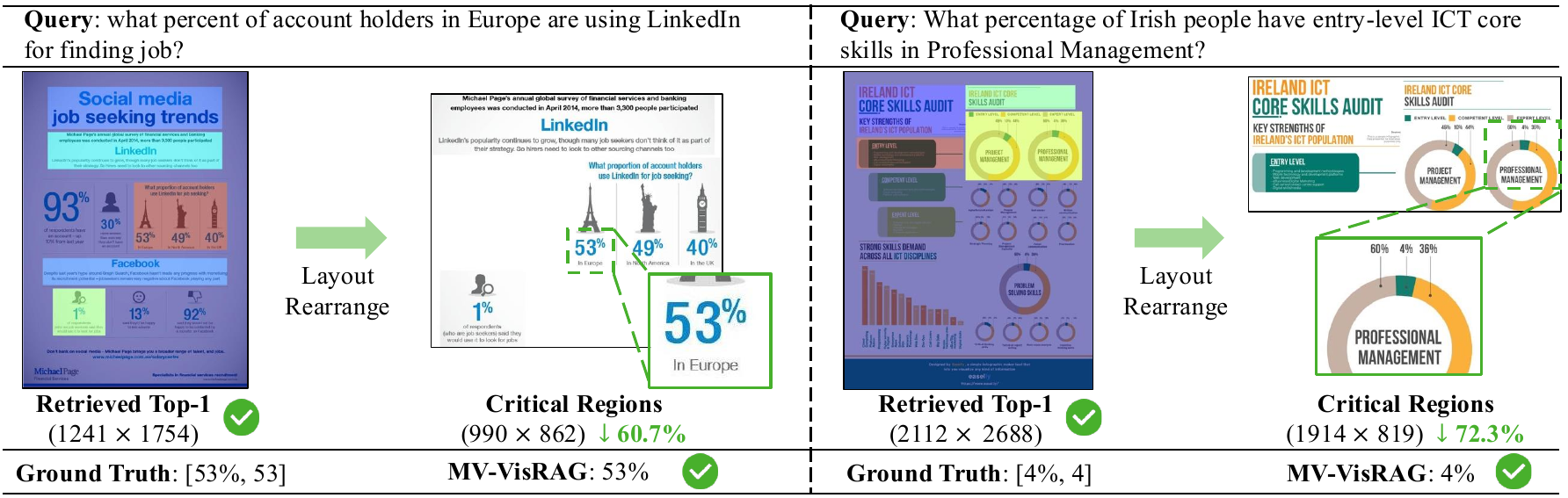}
    \vskip -0.1in
    \caption{Qualitative cases. The heatmap over the document shows the contributions of the corresponding regions.}
    \label{fig-cases}
    \vskip -0.15in
\end{figure*}

\subsection{Generation Results}
\label{exp-generation}
In Table~\ref{tab-generation}, we compare \ourname{} with representative Vision-RAG methods, each of which uses Qwen2.5-VL-3B as the answer generator and the top three retrieved documents as external knowledge. We also evaluate a Text-RAG baseline using the text-only counterpart Qwen2.5~\cite{qwen2025qwen25}, with Tesseract~\cite{smith2007overview} providing OCR text from documents retrieved by our retriever. For \ourname{}, page-level augmentation uses full page images, while region-level augmentation uses selected regions, as described in Section~\ref{sec-selective-rag}.

Both variants of \ourname{} outperform Text-RAG with the same retriever, particularly on InfoVQA, highlighting the value of visual information beyond OCR text. Moreover, our page-level variant beats the other page-level Vision-RAG methods, showing that stronger retrieval benefits generation. As for the region-level variant, it also outperforms its competitor RegionRAG across all benchmarks and metrics.

Compared with our page-level variant, region-level augmentation slightly improves both ANLS and F1 on InfoVQA, as well as F1 on TATDQA. The largest declines occur on DocVQA and DUDE, where our per-query analysis in Appendix~\ref{app-generation_question_subsets} identifies table-related numeric queries as the main source of the drops. Answering such queries often requires row or column headers, which region selection and cropping can omit or separate from the associated values.
In terms of efficiency, Table~\ref{tab-generation-efficiency} shows that region-level augmentation reduces visual tokens by 43\%--62\% relative to page-level augmentation and uses fewer tokens than RegionRAG. The cost analysis in Appendix~\ref{app-storage_runtime_costs} further shows lower generation latency.

\begin{table}[t]
\setlength{\tabcolsep}{4pt}
\caption{Ablation study on key components of the Phi3V-based variant. ``CAR'' means Content-Adaptive Regionalization; disabling CAR replaces it with a fixed $3 \times 3$ grid partition. ``TR'' and ``IR'' denote the text and image reconstruction tasks. }
\vskip -0.1in
\label{tab-retrieval-ablation}
\resizebox{\linewidth}{!}{
\begin{tabular}{ccc|cccc}
\toprule
\multicolumn{3}{c}{\textbf{Components}} & \multicolumn{2}{c}{\textbf{InfoVQA}} & \multicolumn{2}{c}{\textbf{DocVQA}} \\
\cmidrule{1-3}
\cmidrule{4-5}
\cmidrule{6-7}
CAR     & TR     & IR               & nDCG@5                    & Recall@5                  & nDCG@5                    & Recall@5     \\
\midrule
\cmark      & \cmark    & \cmark    & \textbf{73.61}            & \textbf{83.87}            & \textbf{78.96}            & \textbf{88.83} \\
\midrule
\cmark      & \xmark    & \cmark    & \declinecell{69.35}{4.26} & \declinecell{80.15}{3.72} & \declinecell{75.77}{3.19} & \declinecell{84.60}{4.23} \\
\cmark      & \cmark    & \xmark    & \declinecell{71.55}{2.06} & \declinecell{82.06}{1.81} & \declinecell{78.07}{0.89} & \declinecell{86.80}{2.03} \\
\cmark      & \xmark    & \xmark    & \declinecell{70.93}{2.68} & \declinecell{81.58}{2.29} & \declinecell{76.21}{2.75} & \declinecell{85.95}{2.88} \\
\xmark      & \cmark    & \cmark    & \declinecell{70.34}{3.27} & \declinecell{80.87}{3.00} & \declinecell{75.86}{3.10} & \declinecell{85.65}{3.18} \\
\bottomrule
\end{tabular}
}
\vskip -0.15in
\end{table}

\subsection{Ablation Study and Model Discussion}
\label{sec-ablation}

\noindent \textbf{Is Content-Adaptive Regionalization Necessary?}
We replace CAR with fixed $3 \times 3$ grids for region-based multi-vector encoding.
Table~\ref{tab-retrieval-ablation} shows substantial drops in both nDCG@5 and Recall@5, supporting the use of content-adaptive regions for retrieval.
Appendix~\ref{app-car_analysis} provides further evidence that CAR better preserves answer-bearing regions, together with sensitivity and robustness analyses.

\noindent \textbf{Impact of Multimodal Reconstruction Pretraining.}
Table~\ref{tab-retrieval-ablation} ablates the impact of text and image reconstruction on the retriever by setting the corresponding loss weights to zero. 
The results show that removing either of the supervisions degrades performance. Furthermore, removing pretraining (both ``TR'' and ``IR'') leads to average drops of 2.71 nDCG@5 points and 2.58 Recall@5 points.

\begin{figure}[t]
    \centering
    \begin{subfigure}[b]{0.49\linewidth}
        \centering
        \includegraphics[width=\linewidth]{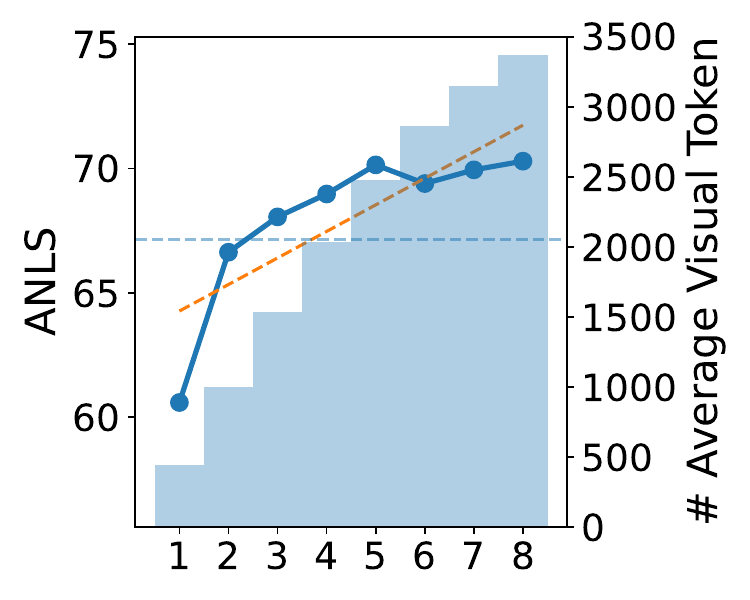}
        \vskip -0.1in
        \caption{Top-K ($K_r$).}
        \label{fig-ablation-topk}
    \end{subfigure}
    \centering
    \begin{subfigure}[b]{0.49\linewidth}
        \centering
        \includegraphics[width=\linewidth]{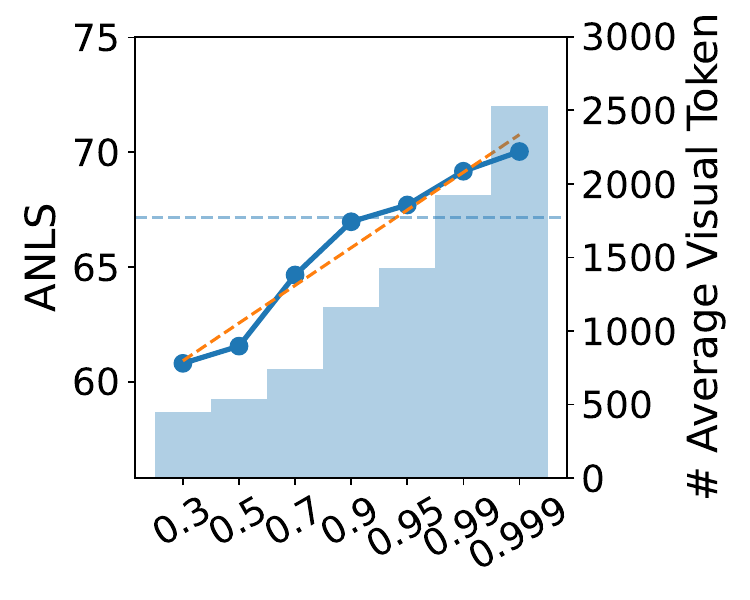}
        \vskip -0.1in
        \caption{Pareto (Threshold $\rho$).}
        \label{fig-ablation-threshold}        
    \end{subfigure}
    \vskip -0.1in
    \caption{Ablation study on region selection strategies.}
    \label{fig-generation-ablation}
    \vskip -0.15in
\end{figure}

\noindent \textbf{Effectiveness of Region Selection.}
Figure~\ref{fig-ablation-topk} examines generation accuracy and visual token counts on InfoVQA under different Top-K settings.
Accuracy improves as $K_r$ increases from 1 to 5, after which additional regions offer limited gains while token counts continue to grow.
Figure~\ref{fig-ablation-threshold} evaluates Pareto-based selection with varying contribution thresholds $\rho$.
Higher thresholds improve ANLS but require more visual tokens. At $\rho=0.999$, this strategy approaches the performance of Top-K with $K_r=5$ at a similar token count.
Together, these results suggest that a small set of high-contribution regions provides sufficient evidence for generation.
We therefore adopt Top-K selection with $K_r=5$ as the default configuration, balancing generation accuracy and visual token cost.

\begin{figure}
    \centering
    \includegraphics[width=\linewidth]{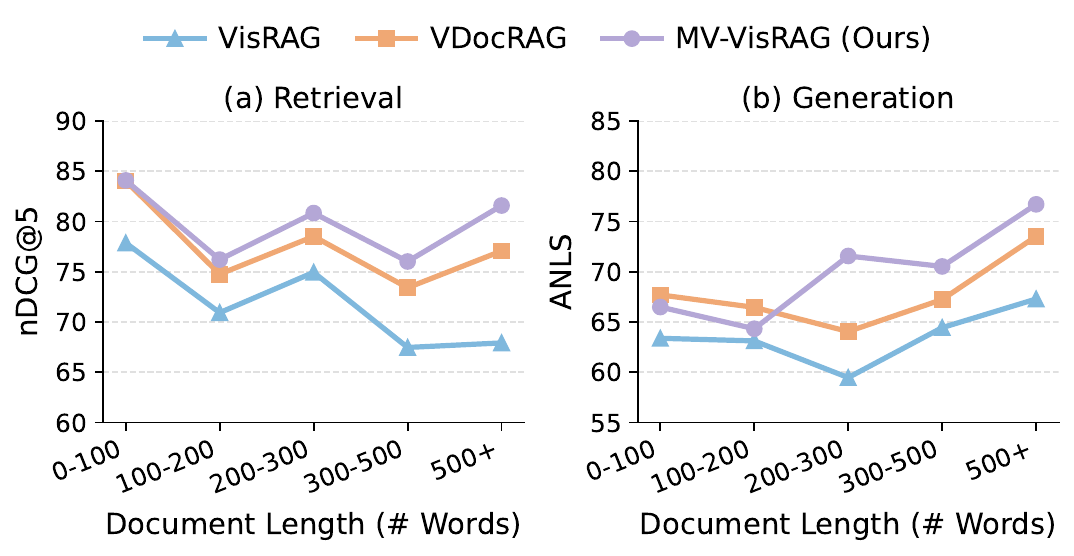}
    \vskip -0.1in
    \caption{Performance with varying document lengths.}
    \label{fig-distribution-to-length}
    \vskip -0.15in
\end{figure}

\noindent \textbf{Performance on Different Document Lengths.}
Figure~\ref{fig-distribution-to-length} compares VisRAG, VDocRAG, and \ourname{} on InfoVQA subsets grouped by the number of OCR-extracted words in the ground-truth documents. All methods use Qwen2.5-VL as the answer generator.
The retrieval advantage of \ourname{} becomes more pronounced on longer documents, suggesting that region-based matching helps locate relevant evidence within text-rich pages.
For generation, \ourname{} outperforms both baselines on the three longer-document subsets, although VDocRAG performs better on the two shorter-document subsets.
Interestingly, all three methods achieve their highest generation scores on the 500+ word subset.
One possible explanation is that these text-rich documents provide more explicit textual cues for answering questions, whereas documents with less text require greater reliance on visual interpretation. This also suggests that effectively leveraging visual evidence for generation remains challenging in Vision-RAG.

\subsection{Qualitative Analysis}
\label{exp-qualitation}
In Figure~\ref{fig-cases}, we visualize the identified critical regions within the retrieved documents, as well as the rearranged region layouts used for region-augmented generation.
These examples show that our method accurately identifies regions relevant to the query while substantially reducing the size of input documents.
Appendix~\ref{app-qualitative_case_studies} also reports some qualitative examples for successful and failed cases.

\section{Conclusion}
In this work, we presented \ourname{}, a Vision-RAG framework that decomposes visually rich documents into semantically complete regions for fine-grained retrieval and generation.
\ourname{} combines a region-based multi-vector retriever with selective region-augmented generation, which selects query-relevant regions based on their contributions to late-interaction similarity and rearranges them into a compact visual context.
The results highlight the value of semantically complete regions as a common basis for document matching and evidence selection, supporting effective retrieval and answer generation with reduced visual input.

\section{Limitations}

Despite the effectiveness of \ourname{}, this work has three main limitations that could be addressed in future research.
(1) The proposed multi-vector retriever is built on multi-image VLMs, whose capacity remains limited when processing a large number of image regions. Consequently, we impose an upper bound of 20 on the number of regions per document. Future work could extend this capacity to better handle documents with more complex layouts.
(2) \ourname{} relies on Content-Adaptive Regionalization (CAR) for document preprocessing, using predefined grouping and merging rules. Region partitioning is therefore not jointly learned with the retriever. A promising direction is to unify region partitioning and encoding within the multi-vector retriever, enabling the learnable identification of semantically complete regions.
(3) Region-level augmentation may miss necessary context distributed across regions. Our analysis on DocVQA and DUDE identifies table-related numeric queries as particularly affected: region selection and cropping can omit row or column headers or separate them from the associated values (Appendix~\ref{app-generation_question_subsets}). Future work could improve region selection by better modeling cross-region dependencies.

\bibliography{main}

\clearpage
\appendix

\section{Implementation Details}
\subsection{Content-Adaptive Regionalization}
\label{app-car}
\paragraph{Text-aware OCR.} 
We apply Tesseract~\cite{smith2007overview} to extract a set of part-level bounding boxes from the document image, each corresponding to the normalized spatial location of a textual part.
Then, we iteratively merge overlapping or adjacent boxes to form coherent text regions. The merging process is shown in Algorithm~\ref{alg-text_aware_ocr}. The distance function $d(\cdot)$ is defined as:
\begin{equation}
d(R_i, R_j)
=
\min_{m,n \in \{1,\dots,4\}}
\left\| \mathbf{c}^{(m)}_i - \mathbf{c}^{(n)}_j \right\|_2 ,
\end{equation}
where $\mathbf{c}^{(m)}_i$ and $\mathbf{c}^{(n)}_j$ denote the $m$-th and $n$-th corner of region $R_i$ and $R_j$, respectively. $\left\| \cdot \right\|_2$ is applied to compute the Euclidean distance between a pair of corners.

\paragraph{Vision-based CCD.}
A Union-Find-based strategy is adopted to efficiently identify visually connected components by clustering spatially adjacent patches with similar visual characteristics. The procedure consists of three main stages, summarized in Algorithm~\ref{alg-vision_ccd}.

\textit{Step 1: Conservative Adjacent Patch Grouping.}
Given an input image $I \in \mathbb{R}^{H \times W \times C}$, the image is first normalized and partitioned into a regular grid of non-overlapping patches 
$\mathcal{P} = \{P_{i,j}\}_{i=1,j=1}^{H', W'}$. 
A local adjacency graph is then constructed by connecting each patch to its spatial neighbors, including the rightward and downward neighbors. 
For each adjacent patch pair, the $\ell_2$ distance between their patch pixels is computed as a visual similarity score. Patch pairs whose distance is below a predefined threshold $\tau_{\text{adj}}$ are merged using a Union-Find data structure. This step produces a conservative set of initial, visually coherent regions, which serve as the basis for subsequent region-level statistics.

\textit{Step 2: Region-level Background Identification.}
For each initial region, all pixels covered by its constituent patches are aggregated to compute the region-level mean intensity and standard deviation. A region is classified as background if its mean intensity exceeds the threshold $\mu_{bg}$ and its standard deviation falls below the threshold $\sigma_{bg}$. This criterion is tailored to document-like inputs, where background regions are typically bright and visually homogeneous. Compared with patch-level detection, region-level statistics provide more stable estimates and are less sensitive to isolated noisy patches.

\begin{algorithm}[t]
\caption{Text-aware OCR}
\label{alg-text_aware_ocr}
\small
\begin{algorithmic}[1]
\Require Image $I$ with width $W$ and height $H$; Part-level bounding boxes $\boldsymbol{B} = \{B_1, \ldots, B_L\}$; Adjacent distance threshold $\tau_{dis}$
\Ensure Text region set $\boldsymbol{R}_{txt}$

\State $\boldsymbol{B} \gets \mathrm{NormalizeBoxes}(\boldsymbol{B}, W, H)$
\State Initialize $\boldsymbol{R}_{txt} \leftarrow \boldsymbol{B}$
\State $\textit{merged} \leftarrow \text{true}$

\While{$\textit{merged}$}
    \State $\textit{merged} \leftarrow \text{false}$
    \ForAll{$(R_i, R_j) \in \boldsymbol{R}_{txt} \times \boldsymbol{R}_{txt},\ i \neq j$}
        \If{$d(R_i, R_j) < \tau_{dis}$}
            \State $R_i \leftarrow R_i \cup R_j$
            \State Remove $R_j$ from $\boldsymbol{R}_{txt}$
            \State $\textit{merged} \leftarrow \text{true}$
            \State \textbf{break}
        \EndIf
    \EndFor
\EndWhile
\State $\boldsymbol{R}_{txt} \gets \mathrm{UnnormalizeBoxes}(\boldsymbol{R}_{txt}, W, H)$
\State \Return $\boldsymbol{R}_{txt}$
\end{algorithmic}
\end{algorithm}

\textit{Step 3: Region-level Foreground Merging.}
After background regions are removed, a second Union-Find procedure is applied at the region level to merge over-fragmented foreground regions. Specifically, two adjacent non-background regions are merged if their region-level mean intensities and standard deviations are sufficiently close, controlled by the merging thresholds $\tau_{merge}^{\mu}$ and $\tau_{merge}^{\sigma}$. This second-stage merging consolidates foreground structures split by the conservative patch grouping in Step~1, while avoiding the direct merging of all adjacent non-background patches. The final output is a set of visual connected components $\boldsymbol{C}_{vis}$, where each component is represented by the indices of its constituent foreground patches.

\begin{algorithm}[]
\caption{Vision-based CCD}
\label{alg-vision_ccd}
\small
\begin{algorithmic}[1]
\Require Image $I \in \mathbb{R}^{H \times W \times C}$; patch size $s$; adjacency threshold $\tau_{\text{adj}}$; background statistics $(\mu_{bg}, \sigma_{bg})$; foreground merging thresholds $(\tau_{merge}^{\mu}, \tau_{merge}^{\sigma})$
\Ensure Visual connected components $\boldsymbol{C}_{vis}$

\State Normalize $I$ and partition it into a regular grid of patches
$\boldsymbol{P} = \{P_{i,j} \in \mathbb{R}^{s \times s \times C}\}_{i=1,j=1}^{H',W'}$.

\State Initialize a Union-Find structure $\mathcal{U}$ over all patches

\State /* Step 1: Conservative Adjacent Patch Grouping */
\ForAll{patch indices $(i,j)$}
    \ForAll{neighbors $(m,n)$ to the right and below of $(i,j)$}
        \If{$\lVert P_{i,j} - P_{m,n} \rVert_2 < \tau_{\text{adj}}$}
            \State $\mathcal{U}.\text{union}((i,j),(m,n))$
        \EndIf
    \EndFor
\EndFor

\State Obtain the initial assignment map
$\boldsymbol{A}^{(0)} \leftarrow \mathcal{U}.\text{find}(\cdot)$

\State Initialize background region set $\mathcal{B} \leftarrow \emptyset$
\State Initialize region statistics dictionary $\mathcal{S} \leftarrow \emptyset$

\State /* Step 2: Region-level Background Identification */
\ForAll{unique labels $a$ in $\boldsymbol{A}^{(0)}$}
    \State Form initial region:
    \Statex \hspace{\algorithmicindent}
    $R_a \leftarrow \{(i,j) \mid \boldsymbol{A}^{(0)}_{i,j}=a\}$

    \State Collect all pixels covered by patches in $R_a$:
    \Statex \hspace{\algorithmicindent}
    $\Omega_a \leftarrow \bigcup_{(i,j) \in R_a} P_{i,j}$

    \State Compute grayscale values:
    \Statex \hspace{\algorithmicindent}
    $x_p \leftarrow \frac{1}{C}\sum_{c=1}^{C} I_p^{(c)}, \quad p \in \Omega_a$

    \State Compute region-level mean and variance:
    \Statex \hspace{\algorithmicindent}
    $\mu_a \leftarrow \frac{1}{|\Omega_a|}\sum_{p \in \Omega_a} x_p$
    \Statex \hspace{\algorithmicindent}
    $\sigma_a \leftarrow 
    \sqrt{
    \frac{1}{|\Omega_a|}
    \sum_{p \in \Omega_a}(x_p-\mu_a)^2
    }$

    \State Store region statistics:
    \Statex \hspace{\algorithmicindent}
    $\mathcal{S}[a] \leftarrow (\mu_a, \sigma_a)$

    \If{$\mu_a > \mu_{bg}$ \textbf{and} $\sigma_a < \sigma_{bg}$}
        \State $\mathcal{B} \leftarrow \mathcal{B} \cup \{a\}$
    \EndIf
\EndFor

\State /* Step 3: Region-level Foreground Merging */
\State Initialize a Union-Find structure $\mathcal{U}_R$ over all non-background region labels

\ForAll{neighboring patch pairs $((i,j),(m,n))$}
    \State $a \leftarrow \boldsymbol{A}^{(0)}_{i,j}$, $b \leftarrow \boldsymbol{A}^{(0)}_{m,n}$

    \If{$a \neq b$ \textbf{and} $a \notin \mathcal{B}$ \textbf{and} $b \notin \mathcal{B}$}
        \State Get region statistics: 
        \State $(\mu_a,\sigma_a) \leftarrow \mathcal{S}[a]$, $(\mu_b,\sigma_b) \leftarrow \mathcal{S}[b]$

        \If{$|\mu_a-\mu_b| < \tau_{merge}^{\mu}$ 
        \textbf{and} $|\sigma_a-\sigma_b| < \tau_{merge}^{\sigma}$}
            \State $\mathcal{U}_R.\text{union}(a,b)$
        \EndIf
    \EndIf
\EndFor

\State /* Step 4: Connected Component Extraction */
\State Initialize $\boldsymbol{C}_{vis} \leftarrow \emptyset$

\ForAll{unique labels $q$ in $\mathcal{U}_R.\text{find}(\cdot)$}
    \State Form merged visual component:
    \Statex \hspace{\algorithmicindent}
    $C_q \leftarrow 
    \{(i,j) \mid 
    \boldsymbol{A}^{(0)}_{i,j} \notin \mathcal{B}
    \textbf{ and }
    \mathcal{U}_R.\text{find}(\boldsymbol{A}^{(0)}_{i,j}) = q
    \}$

    \If{$C_q \neq \emptyset$}
        \State $\boldsymbol{C}_{vis} \leftarrow \boldsymbol{C}_{vis} \cup \{C_q\}$
    \EndIf
\EndFor

\State \Return $\boldsymbol{C}_{vis}$
\end{algorithmic}
\end{algorithm}

\paragraph{Finalization with Region Merging.}
To integrate textual and visual evidence, we merge textual regions $\boldsymbol{R}_{txt}$ with visual connected components $\boldsymbol{C}_{vis}$. 
Each visual component $C_v \in \boldsymbol{C}_{vis}$ is first converted into a vision region by taking the bounding-box union of its constituent patches, yielding a set of vision regions $\boldsymbol{R}_{vis}$. 
For each vision region $R_v \in \boldsymbol{R}_{vis}$, we identify the textual region with the highest minimum-area overlap score:
\begin{equation}
R_t^{*}
=
\arg\max_{R_t \in \boldsymbol{R}_{txt}}
\mathrm{IoM}(R_v, R_t).
\end{equation}
The two regions are merged if their overlap score satisfies $\mathrm{IoM}(R_v, R_t^{*}) > \tau_{merge}$, where $\mathrm{IoM}$ is defined as
\begin{equation}
\mathrm{IoM}(R_i, R_j)
=
\frac{\left| R_i \cap R_j \right|}
{\min\!\left( \left| R_i \right|,\; \left| R_j \right| \right)} .
\end{equation}
This criterion measures the extent to which the smaller region is covered by the larger one, making it suitable for merging nested or highly overlapping textual and visual regions.

After merging, oversized regions whose heights exceed a predefined threshold $\tau_{height}$ are split along the height ($y$) axis into horizontal bands, and regions with an area smaller than $\tau_{area}$ are removed as isolated noisy fragments. 
Both height and area are measured in the original pixel space, with $\tau_{height}$ and $\tau_{area}$ determined by the patch size $s$, as summarized in Table~\ref{tab-hyperparameters}. 
The refinement produces a set of extended regions $\boldsymbol{R} = \{R_1, \ldots, R_K\}$ that serve as spatially coherent units, jointly incorporating textual and visual cues.

\subsection{Region Rearrangement}
\label{app-region-rearrangement}

\begin{figure*}
    \centering
    \includegraphics[width=\linewidth]{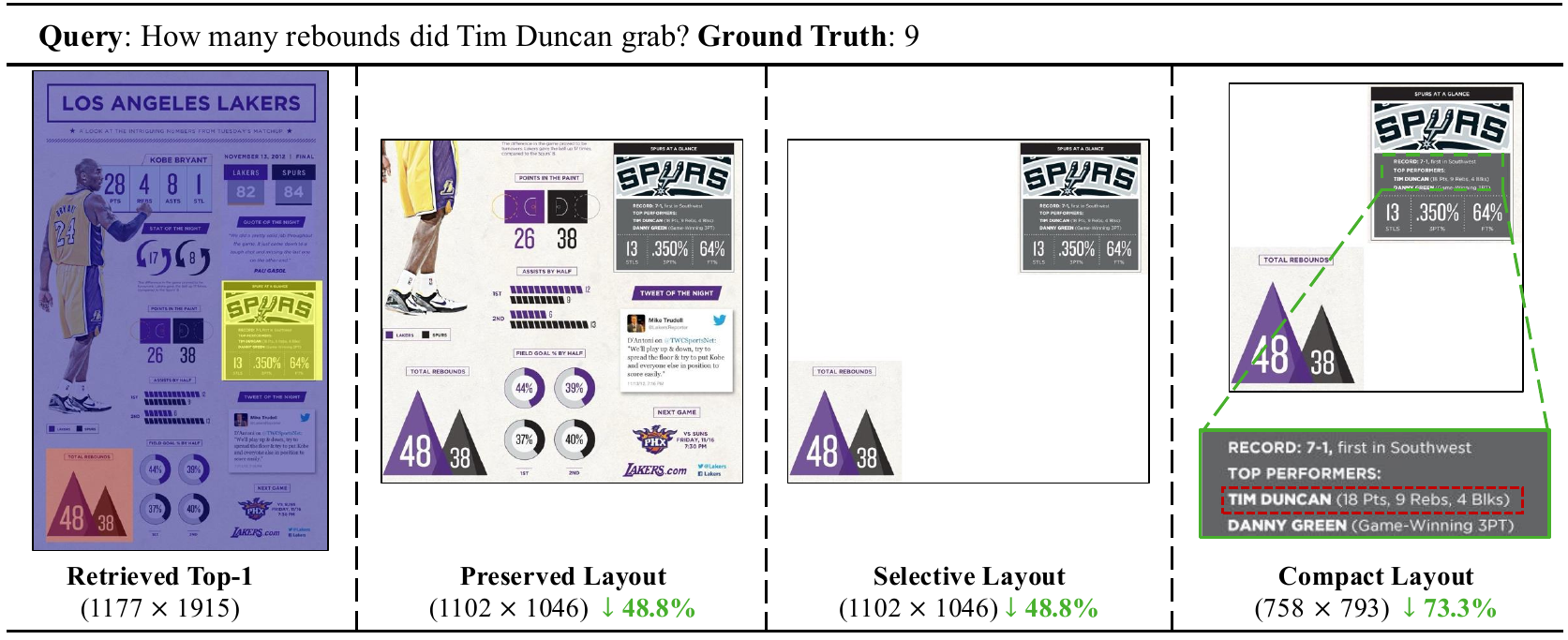}
    \caption{Different layout rearrangement strategies.}
    \label{fig-layouts}
\end{figure*}

To enable region-level visual augmentation, we investigate three strategies for rearranging critical regions: Preserved Layout, Selective Layout, and Compact Layout, as illustrated in Figure~\ref{fig-layouts}.

Let $\boldsymbol{R}_{critical} = \{R_i\}_{i=1}^{K_r}$ denote the set of critical regions, where each region $R_i$ is represented by a bounding box $(x_1^i, y_1^i, x_2^i, y_2^i)$. Here, $(x_1^i, y_1^i)$ and $(x_2^i, y_2^i)$ correspond to the top-left and bottom-right corners of the region, respectively.

\noindent \textbf{(1) Preserved Layout} preserves the local spatial context within the bounding extent of the critical regions. Specifically, it constructs a single crop spanning from the minimal top-left corner to the maximal bottom-right corner among all critical regions:
$(\min_i x_1^i, \min_i y_1^i, \max_i x_2^i, \max_i y_2^i)$.
This strategy retains both the critical regions and the intermediate visual content enclosed by their extreme coordinates.

\noindent \textbf{(2) Selective Layout} uses the same spatial extent as Preserved Layout but keeps only the visual content inside the critical regions, replacing all other areas with whitespace. This design preserves the original global arrangement of the selected regions while suppressing irrelevant surrounding content.

\noindent \textbf{(3) Compact Layout} further reduces redundant whitespace by rearranging critical regions into a more compact canvas. The regions are pasted without resizing, preserving their original visual content and scale. Their absolute positions and long-range distances are discarded, while axis-wise relative ordering is maintained.
The implementation details of Compact Layout are illustrated in Algorithm~\ref{alg-compact_layout}. First, all regions are projected onto the horizontal and vertical axes to form one-dimensional intervals, which are sorted and merged into axis-wise content blocks (Lines 1--6). 
Next, the content blocks are compacted independently along each axis by sequentially relocating them toward the origin while preserving a bounded inter-block margin. The corresponding displacement of each block is recorded, and the compacted canvas size is determined (Lines 7--20).
Finally, a blank canvas is initialized, and each region is translated by the horizontal and vertical shifts of the blocks that contain it. The region is then cropped from the original image and pasted onto the compacted canvas without resizing (Lines 21--31).
This axis-decoupled strategy removes large horizontal and vertical whitespace gaps while preserving the visual content and local structure of the selected regions.

Through empirical evaluation, we observe that Compact Layout substantially reduces redundant canvas area while maintaining nearly lossless VQA accuracy. \textbf{Hence, we adopt Compact Layout as the default region rearrangement strategy.} A detailed experimental analysis is provided in Section~\ref{app-layout-ablation}.

\begin{algorithm}[t]
\caption{Compact Layout Rearrangement}
\label{alg-compact_layout}
\small
\begin{algorithmic}[1]
\Require Image $I$; region set $\boldsymbol{R}=\{(x_1,y_1,x_2,y_2)\}$; inter-block margin $m$; canvas padding $p$
\Ensure Compacted image $\hat{I}$

\State /* Step 1: Axis-wise Interval Extraction */
\For{each axis $a \in \{x,y\}$}
    \State Extract 1D intervals $I_a$ from $\boldsymbol{R}$ along axis $a$
    \State Sort intervals in $I_a$ by their start coordinates
    \State Merge overlapping intervals to obtain ordered content blocks $C_a$
\EndFor

\State /* Step 2: Block Compaction and Shift Computation */
\For{each axis $a \in \{x,y\}$}
    \State Initialize cursor position $c \leftarrow 0$
    \For{each block $(s,e) \in C_a$ in order}
        \State Compute block length $\ell \leftarrow e - s$
        \State Assign block shift $\Delta_a(s,e) \leftarrow c - s$
        \State Update cursor $c \leftarrow c + \ell$
        \If{next block $(s_{\text{next}}, e_{\text{next}})$ exists}
            \State Original inter-block gap $g \leftarrow s_{\text{next}} - e$
            \State Add bounded margin $c \leftarrow c + \min(m, g)$
        \EndIf
    \EndFor
    \State Record total compacted size $S_a \leftarrow c$
\EndFor

\State /* Step 3: Canvas Initialization */
\State Create a blank canvas $\hat{I}$ with size ($S_x + 2p$, $S_y + 2p$)

\State /* Step 4: Region Relocation */
\For{each region $(x_1,y_1,x_2,y_2) \in \boldsymbol{R}$}
    \State Find horizontal block $(s_x,e_x) \in C_x$ containing $x_1$
    \State Find vertical block $(s_y,e_y) \in C_y$ containing $y_1$
    \State Get shifts $\Delta_x \leftarrow \Delta_x(s_x,e_x)$ and $\Delta_y \leftarrow \Delta_y(s_y,e_y)$
    \State Compute new coordinates:
    \Statex \hspace{\algorithmicindent}
    $\hat{x}_1 \leftarrow x_1 + \Delta_x + p,\quad
    \hat{y}_1 \leftarrow y_1 + \Delta_y + p$
    \Statex \hspace{\algorithmicindent}
    $\hat{x}_2 \leftarrow x_2 + \Delta_x + p,\quad
    \hat{y}_2 \leftarrow y_2 + \Delta_y + p$
    \State Crop region $I[y_1:y_2, x_1:x_2]$ from $I$
    \State Paste the cropped region into $\hat{I}$ at $(\hat{x}_1,\hat{y}_1,\hat{x}_2,\hat{y}_2)$
\EndFor

\State \Return $\hat{I}$
\end{algorithmic}
\end{algorithm}

\begin{table*}[t]
\centering
\caption{Model checkpoints and licenses on Hugging Face.}
\label{tab-baseline}
\resizebox{0.95\linewidth}{!}{
\begin{tabular}{lll}
\toprule
\textbf{Model} & \textbf{Model Checkpoint} & \textbf{License} \\
 \midrule
 Contriever    & \url{https://huggingface.co/facebook/contriever-msmarco} & Not specified \\
 E5            & \url{https://huggingface.co/intfloat/e5-base-v2} & MIT \\
 GTE           & \url{https://huggingface.co/thenlper/gte-base} & MIT \\
 E5-Mistral    & \url{https://huggingface.co/intfloat/e5-mistral-7b-instruct} & MIT \\
 NV-Embed-v2   & \url{https://huggingface.co/nvidia/NV-Embed-v2} & CC-BY-NC-4.0 \\
 DSE           & \url{https://huggingface.co/Tevatron/dse-phi3-docmatix-v1} & MIT \\
 UniME-V2      & \url{https://huggingface.co/TianchengGu/UniME-V2-Qwen2VL-7B} & Apache-2.0 \\
 VLM2Vec-V2    & \url{https://huggingface.co/VLM2Vec/VLM2Vec-V2.0} & Apache-2.0 \\
 VisRAG        & \url{https://huggingface.co/openbmb/VisRAG-Ret} & Apache-2.0 \\
 VDocRAG (Phi3V) & \url{https://huggingface.co/NTT-hil-insight/VDocRetriever-Phi3-vision} & NTT License \\
 RegionRAG     & \url{https://huggingface.co/Aeryn666/RegionRet} & MIT \\
 Phi3V         & \url{https://huggingface.co/microsoft/Phi-3-vision-128k-instruct} & MIT \\
 Qwen2.5       & \url{https://huggingface.co/Qwen/Qwen2.5-3B-Instruct} & Apache-2.0 \\
 Qwen2.5-VL    & \url{https://huggingface.co/Qwen/Qwen2.5-VL-3B-Instruct} & Apache-2.0 \\
 \bottomrule
\end{tabular}
}
\end{table*}

\section{Experimental Details}
\label{app-experimental-details}
\subsection{Baselines}
To ensure a fair comparison with representative single-vector and multi-vector methods, we reproduce VDocRAG~\cite{tanaka2025vdocrag} and ColQwen~\cite{faysse2025colpali} using their official implementations, with the same Qwen2.5-VL~\cite{bai2025qwen25vl} backbone and OpenDocVQA~\cite{tanaka2025vdocrag} fine-tuning data as our method.
DocPruner~\cite{yan2025docpruner}, Light-ColPali~\cite{ma2025towards}, and PRUNE-THEN-MERGE~\cite{yan2026sculpting} perform post-hoc pruning or compression on the document embeddings produced by this ColQwen model. Since their code was not publicly available, we implement these methods following the descriptions in their respective papers.

For RegionRAG~\cite{li2026regionrag}, we use the official implementation and the released RegionRet checkpoint without retraining.
Retrieval is evaluated on the same benchmark splits and candidate document pools as the other retrievers.
For generation, RegionRAG uses Qwen2.5-VL-3B with the top three retrieved documents and is evaluated using ANLS and F1, whereas the original RegionRAG paper reports relaxed answer accuracy under top-1/top-4 retrieval settings.

For the remaining baselines, we use publicly released checkpoints.
Table~\ref{tab-baseline} lists the checkpoint URLs and licenses.

\begin{table*}[t]
\centering
\caption{Evaluation dataset statistics.}
\label{tab-benchmark}
\resizebox{\linewidth}{!}{
\begin{tabular}{lrrl}
\toprule
\textbf{Benchmark} & \textbf{\# Queries} & \textbf{\# Documents} & \textbf{Description}                             \\
\midrule
InfoVQA~\cite{mathew2022infographicvqa} & 1,048      & 5,485        & Infographics scraped from the web                         \\
DUDE~\cite{van2023document}             & 496        & 27,955       & Multi-page, multi-domain, and multi-industry long documents \\
DocVQA~\cite{mathew2021docvqa}          & 591        & 741          & Scanned documents from UCSF Industry                       \\
TATDQA~\cite{zhu2022towards}            & 1,662      & 277          & High-quality financial reports                             \\
ViDoRe-P~\cite{faysse2025colpali}         & 400        & 4,000        & Documents about energy, administration, medicine, and AI domains \\
\bottomrule
\end{tabular}
}
\end{table*}

\subsection{Benchmark}
Table~\ref{tab-benchmark} summarizes the evaluation datasets, including query counts, document counts, and document types.
We remove duplicate queries and documents from the original datasets.
The main retrieval experiments use a separate candidate document pool for each benchmark.

InfoVQA, DUDE, and DocVQA share dataset sources with the OpenDocVQA fine-tuning data and are used for in-domain evaluation.
TATDQA and ViDoRe-P are used for out-of-domain evaluation because their dataset sources are not included in fine-tuning.
We further check train--evaluation overlap through exact matching of normalized query texts and document names against the OpenDocVQA fine-tuning data. No overlap is found for either query texts or document names.

\begin{table}[t]
\centering
\caption{Hyperparameters used in our approach.}
\label{tab-hyperparameters}
\resizebox{\linewidth}{!}{
\begin{tabular}{lc}
\toprule
\textbf{Hyperparameter} & \textbf{Value} \\
\midrule
\rowcolor{gray!15} \multicolumn{2}{c}{\textit{\textbf{Content-Adaptive Regionalization}}} \\
distance threshold $\tau_{dis}$ & 0.03 \\ 
adjacency threshold $\tau_{adj}$ & 5.0 \\
background statistics ($\mu_{bg}$, $\sigma_{bg}$) & (0.95, 0.10) \\
foreground merging thresholds $(\tau_{merge}^{\mu}, \tau_{merge}^{\sigma})$ & (0.10, 0.05) \\
region merging threshold $\tau_{merge}$ & $0.3$ \\
patch size $s$ & 28 pixels \\
maximum region height $\tau_{height}$ & $14 \times s$ \\
minimal region area $\tau_{area}$ & $50 \times s^{2} $ \\
\midrule
\rowcolor{gray!15} \multicolumn{2}{c}{\textit{\textbf{Region-based Multi-Vector Retriever}}} \\
loss weighting factor $\alpha$ & 1.0 \\
learning rate & $1\times10^{-4}$ \\
accumulation batch size & 64 \\
AdamW $\beta_1$ & 0.9 \\
AdamW $\beta_2$ & 0.999 \\
LoRA rank r     & 8 \\
LoRA scaling alpha & 64 \\
temperature $\tau $& 0.01 \\
\midrule
\rowcolor{gray!15} \multicolumn{2}{c}{\textit{\textbf{Selective Region-Augmented Generation }}} \\
weighting factor $\beta$ & 0.1 \\
retrieved document number $K_d$ for RAG  & 3 \\
region number $K_r$ & 5 \\
cumulative contribution threshold $\rho$ & 0.99 \\
inter-block margin $m$ & 30 pixels \\
canvas padding $p$ & 0 pixel \\
\bottomrule
\end{tabular}
}
\end{table}

\subsection{Implementation Details.}
For the retriever of \ourname{}, only the LLM components are fine-tuned using low-rank adapters~\cite{hu2022lora} with rank $r=8$ and scaling factor $\alpha=64$, while all other parameters remain frozen.
The accumulation batch size (with gradient accumulation) is 64 for both pretraining and fine-tuning.
The pretraining stage is optimized for one epoch, and the fine-tuning stage for two epochs; both use a learning rate of $1\times10^{-4}$ with linear decay and a 10\% warmup schedule. During both the training and inference phases, the retriever encodes at most 20 regions per document, truncating by the input order of the regions.
The generator of \ourname{} is the off-the-shelf Qwen2.5-VL-3B, used without additional fine-tuning.
For retrieval-augmented generation, the top-3 retrieved documents are used as additional context for all models.
All generation evaluations use deterministic decoding with decoding temperature set to 0 and sampling disabled (\texttt{do\_sample=False}).
In region-level augmentation, the Top-K strategy is adopted by default, selecting up to five critical regions per document.

Detailed hyperparameter settings are listed in Table~\ref{tab-hyperparameters}, covering the Content-Adaptive Regionalization, Region-based Multi-Vector Retriever, and Selective Region-Augmented Generation modules.

\section{Analysis of Content-Adaptive Regionalization}
\label{app-car_analysis}

\begin{figure}[t]
    \centering
    \includegraphics[width=\linewidth]{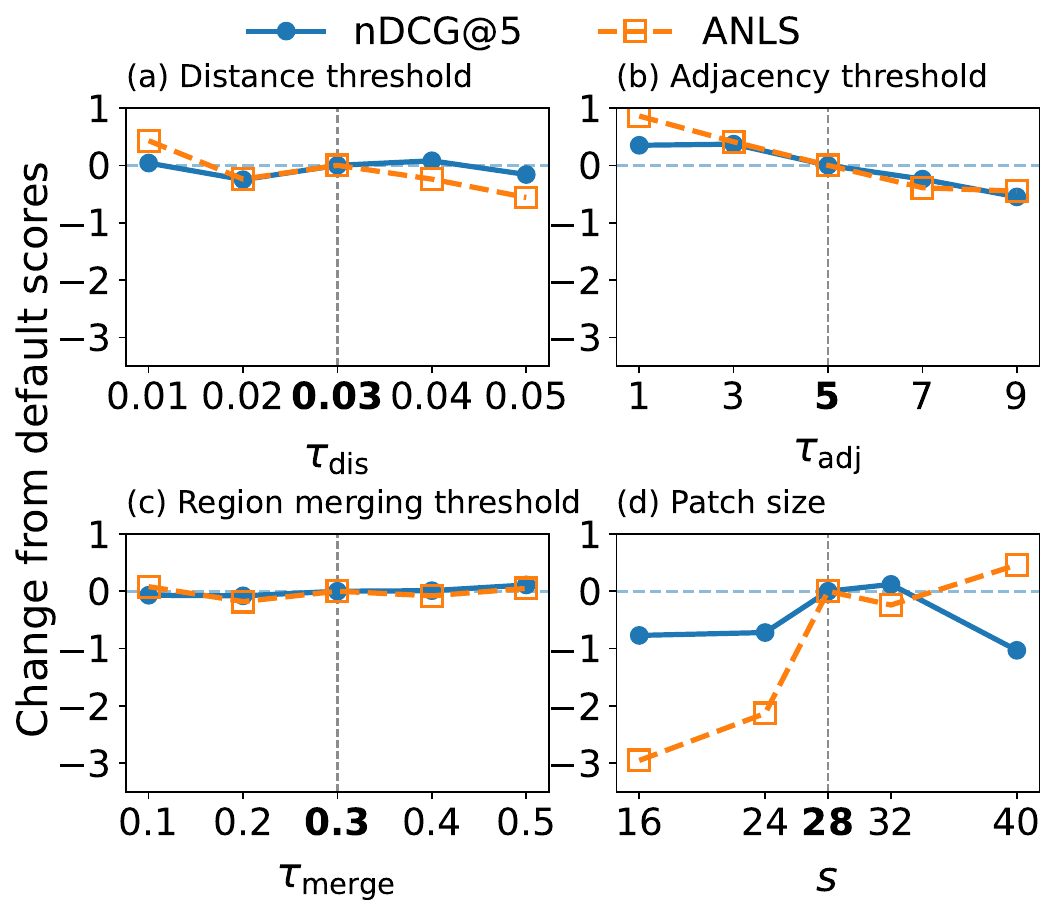}
    \caption{Sensitivity to CAR hyperparameters on InfoVQA. Curves show changes in score points relative to the default configuration, which achieves 79.36 nDCG@5 and 69.34 ANLS. Vertical dashed lines and bold ticks indicate the default parameter values.}
    \label{fig:car_hyperparameter_sensitivity}
\end{figure}

\begin{figure*}[t]
    \centering
    \begin{subfigure}[t]{0.49\textwidth}
        \centering
        \includegraphics[width=\linewidth]{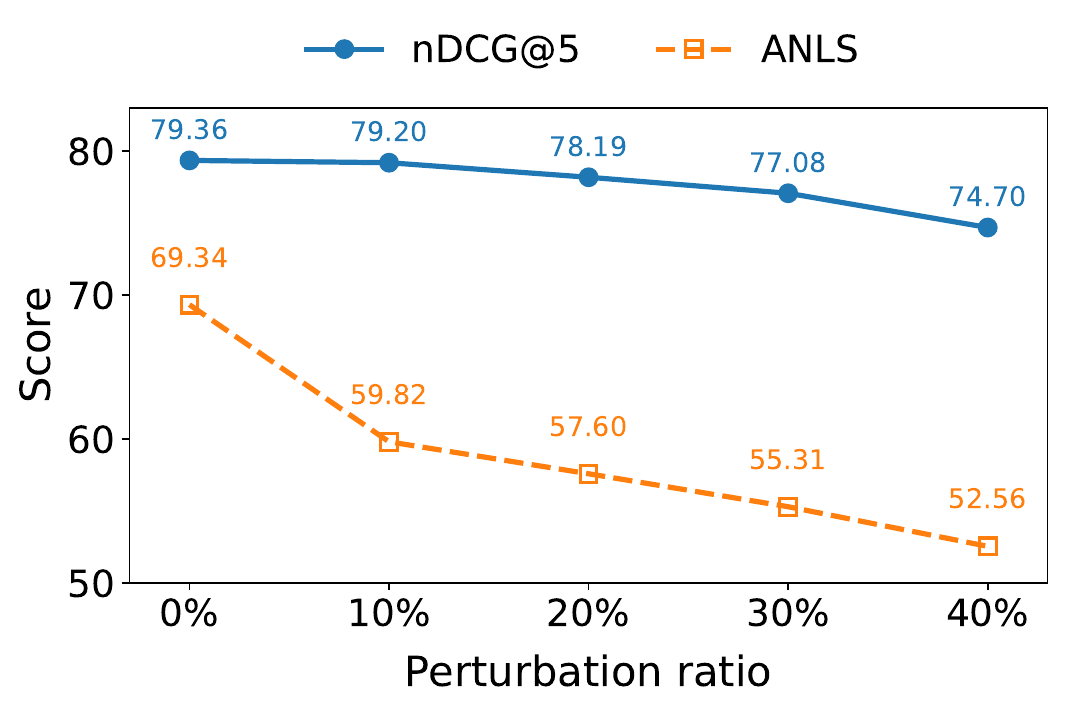}
        \caption{Boundary perturbations.}
        \label{fig:car_shift_robustness}
    \end{subfigure}
    \hfill
    \begin{subfigure}[t]{0.49\textwidth}
        \centering
        \includegraphics[width=\linewidth]{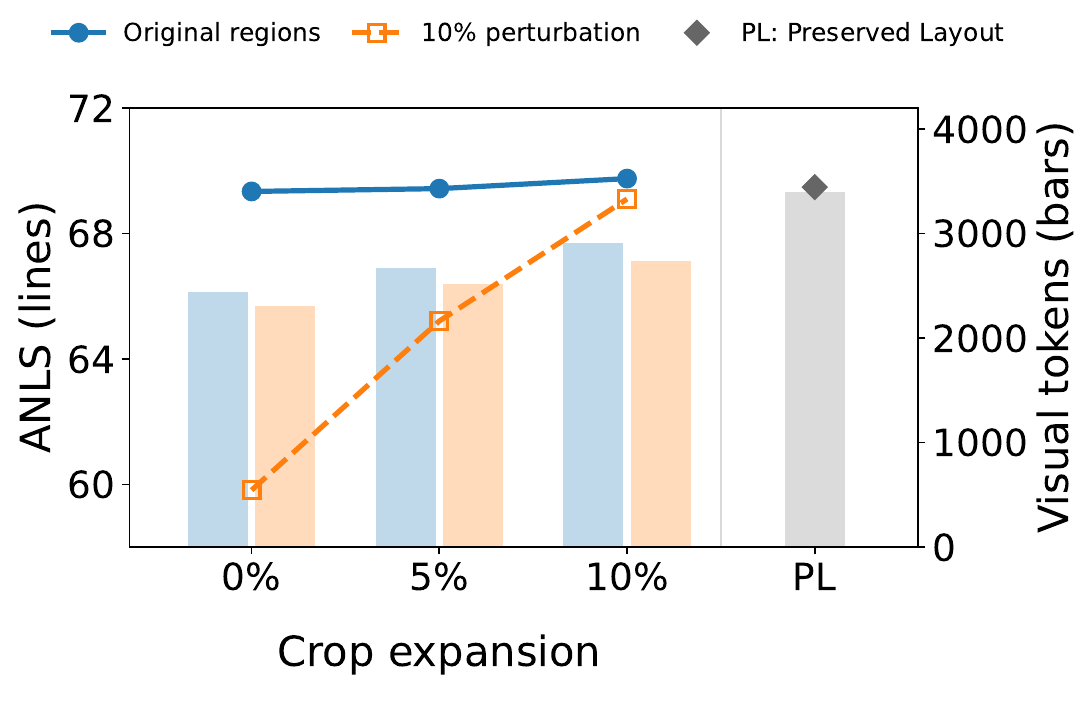}
        \caption{Context expansion.}
        \label{fig:car_context_expansion}
    \end{subfigure}
    \caption{Robustness to region-boundary perturbations and the effect of crop expansion on InfoVQA. (a) Retrieval and generation scores under increasing boundary perturbations. (b) ANLS (lines, left axis) and average visual token counts (bars, right axis) with crop expansion before Compact Layout. PL denotes Preserved Layout using the original CAR regions.}
    \label{fig:car_robustness}
\end{figure*}

\subsection{Sensitivity to CAR Hyperparameters}
\label{app-car_hyperparameter_sensitivity}

CAR's grouping thresholds and patch size determine the granularity of the regions used for retrieval and generation.
We evaluate their effects on InfoVQA by varying the distance threshold $\tau_{dis}$, adjacency threshold $\tau_{adj}$, region merging threshold $\tau_{merge}$, and patch size $s$ individually.
We keep the trained retriever fixed and use the default values in Table~\ref{tab-hyperparameters} for the other independent CAR parameters.
When varying $s$, the maximum region height and minimum region area change accordingly as $\tau_{height}=14s$ and $\tau_{area}=50s^2$.
Generation uses Qwen2.5-VL-3B with region-level augmentation and Compact Layout.

Figure~\ref{fig:car_hyperparameter_sensitivity} shows that retrieval varies little across the tested settings, with the nDCG@5 range within each sweep at most 1.15 points.
The three grouping thresholds also have a small effect on generation, with ANLS ranges of 0.99, 1.31, and 0.26 points for $\tau_{dis}$, $\tau_{adj}$, and $\tau_{merge}$, respectively.
Generation is more sensitive to patch size: ANLS ranges from 66.39 at $s=16$ to 69.80 at $s=40$, a difference of 3.41 points.
We attribute this sensitivity to the coupled effect of $s$ on visual grouping, region splitting, and area filtering. Smaller values can spread answer evidence and its surrounding context across smaller region fragments, so selecting only five regions per document may provide the generator with less complete evidence.

\subsection{Robustness of Content-Adaptive Regionalization}
\label{app-car_robustness}

Region partitioning affects both the content encoded for retrieval and the visual evidence provided to the generator.
We simulate partitioning errors by randomly shifting CAR-generated region bounding boxes by ratios from 0\% to 40\% relative to their original sizes, with 0\% corresponding to the original CAR output.
We evaluate retrieval and generation on InfoVQA, using Qwen2.5-VL-3B with region-level visual augmentation and Compact Layout for generation.

Figure~\ref{fig:car_shift_robustness} shows that retrieval degrades gradually: nDCG@5 decreases from 79.36 to 79.20 at 10\% perturbation and to 74.70 at 40\%.
Generation is more sensitive, with ANLS falling from 69.34 to 59.82 and 52.56 at the same perturbation levels.
These trends suggest that retrieval can still match relevant content within shifted regions, whereas generation is more affected when crops omit answer details or their surrounding context.

To examine whether additional local context can recover generation performance, we expand each selected crop to include surrounding page content before arranging it with Compact Layout.
Each side is extended outward by 0\%, 5\%, or 10\% of the crop's width for horizontal expansion and height for vertical expansion.
We evaluate both the original CAR regions and regions with 10\% boundary perturbation, and report ANLS together with the average visual token count.

As shown in Figure~\ref{fig:car_context_expansion}, under 10\% boundary perturbation, 5\% crop expansion raises ANLS from 59.82 to 65.21, and 10\% crop expansion raises it to 69.10, close to the unperturbed score of 69.34.
The 10\% crop expansion increases the average visual token count from 2,306 to 2,739.
For the original regions, the same expansion ratio improves ANLS from 69.34 to 69.75 while increasing tokens from 2,436 to 2,905, compared with 69.48 ANLS and 3,392 tokens for Preserved Layout.
These results suggest that including neighboring content can recover evidence lost at perturbed boundaries, with additional visual tokens providing more complete context for generation.

\begin{table*}[t]
\centering
\caption{Evaluation on ViDoRe V2. We report nDCG@5 and R@5. The best results are highlighted in bold.}
\label{tab:vidore_v2}
\resizebox{\linewidth}{!}{
\begin{tabular}{lcccccccccc}
\toprule
\multirow{2}{*}{\textbf{Method}} 
& \multicolumn{2}{c}{\textbf{ESG Reports}} 
& \multicolumn{2}{c}{\textbf{Economics}} 
& \multicolumn{2}{c}{\textbf{Biomedical}}
& \multicolumn{2}{c}{\textbf{ESG Reports (Manual)}}
& \multicolumn{2}{c}{\textbf{Average}} \\
\cmidrule(lr){2-3} \cmidrule(lr){4-5} \cmidrule(lr){6-7} \cmidrule(lr){8-9} \cmidrule(lr){10-11}
& nDCG@5 & R@5 & nDCG@5 & R@5 & nDCG@5 & R@5 & nDCG@5 & R@5 & nDCG@5 & R@5 \\
\midrule
VDocRAG            & 36.05 & 42.93 & 50.72 & \textbf{27.99} & 51.00 & 56.76 & 45.03 & 51.04 & 45.70 & 44.68 \\
ColQwen-Pooling    & 44.53 & 47.91 & 47.03 & 25.54 & 54.74 & 58.06 & 49.98 & 59.53 & 49.07 & 47.76 \\
DocPruner          & \textbf{47.69} & 51.34 & 46.22 & 24.72 & 52.88 & 57.33 & 51.02 & 60.66 & 49.45 & 48.51 \\
Light-ColPali      & 44.96 & 50.14 & 42.99 & 20.69 & 55.48 & 58.79 & 49.41 & 56.81 & 48.21 & 46.61 \\
PRUNE-THEN-MERGE   & 46.93 & 51.10 & 46.77 & 25.69 & 54.80 & 58.95 & \textbf{54.71} & \textbf{63.35} & 50.80 & 49.77 \\
\ourname{} (Ours)  & 47.38 & \textbf{51.48} & \textbf{51.40} & 27.82 & \textbf{56.19} & \textbf{59.33} & 51.31 & 60.61 & \textbf{51.57} & \textbf{49.81} \\
\bottomrule
\end{tabular}
}
\end{table*}

\begin{table*}[t]
\centering
\caption{Retrieval performance of \ourname{} using representations at different levels. The results are reported in nDCG@5 and Recall@5.}
\label{tab-local-global}
\resizebox{\textwidth}{!}{
\begin{tabular}{ccccccccccc}
\toprule
 \multirow{2}{*}{\textbf{Embedding}} & \multicolumn{2}{c}{\textbf{InfoVQA}} & \multicolumn{2}{c}{\textbf{DUDE}} & \multicolumn{2}{c}{\textbf{DocVQA}} & \multicolumn{2}{c}{\textbf{TATDQA}} & \multicolumn{2}{c}{\textbf{ViDoRe-P}} \\
 & nDCG@5 & R@5 & nDCG@5 & R@5 & nDCG@5 & R@5 & nDCG@5 & R@5 & nDCG@5 & R@5 \\
\midrule
Global (\eos{})       & 52.57 & 64.03 & 32.82 & 41.43 & 52.92 & 64.47 & 24.99 & 33.01 & 65.59 & 79.75 \\
Local (\eoi{})        & 78.25 & 87.98 & 65.94 & 74.88 & 84.04 & 90.86 & 61.14 & 72.94 & 90.35 & 95.25 \\
All (\eos{} + \eoi{}) & \textbf{79.36} & \textbf{88.65} & \textbf{70.55} & \textbf{79.76} & \textbf{86.20} & \textbf{93.05} & \textbf{61.37} & \textbf{73.18} & \textbf{92.48} & \textbf{96.75} \\
\bottomrule
\end{tabular}
}
\end{table*}

\section{Additional Results}

\subsection{Evaluation on ViDoRe V2}
\label{app-vidore_v2}

To further evaluate the generalization ability of \ourname{}, we conduct additional experiments on the more recent ViDoRe V2~\cite{mace2025vidore} benchmark. Following our main retrieval evaluation, we compare \ourname{} with single-vector and pruning- or merging-based efficient patch-level multi-vector baselines, and report nDCG@5 and R@5 across its ESG Reports, Economics, Biomedical, and ESG Reports (Manual) subsets.

As shown in Table~\ref{tab:vidore_v2}, \ourname{} achieves the best average performance among the evaluated single-vector and efficient multi-vector baselines, reaching 51.57 nDCG@5 and 49.81 R@5. It obtains the highest nDCG@5 on Economics and Biomedical, and the highest R@5 on ESG Reports and Biomedical. Although PRUNE-THEN-MERGE performs better on ESG Reports (Manual), and DocPruner slightly leads on ESG Reports nDCG@5, \ourname{} remains competitive across all subsets. These results are consistent with the main retrieval results: semantically complete region vectors provide a strong and robust alternative to both single-vector retrieval and efficient patch-level multi-vector compression strategies.

\subsection{Effectiveness of Local and Global Embedding in Retrieval}

To investigate the contribution of different visual granularity levels in our multi-vector retriever, we perform a detailed ablation study on the embedding components. Specifically, we compare using only global embeddings (\eos{}) versus only local embeddings (\eoi{}), as shown in Table~\ref{tab-local-global}.
Global embeddings alone show limited retrieval capability, as a single vector compresses the entire document into a coarse representation, discarding fine-grained semantic and structural cues. In contrast, local embeddings consistently yield substantial gains across all datasets. By representing documents as sets of semantically complete regions, local embeddings preserve detailed textual information, enabling more precise matching between queries and relevant regions. 
Switching from global to local embeddings improves nDCG@5 by 30.2 points and Recall@5 by 27.8 points on average, highlighting the importance of region-level representations for our proposed retriever.
Combining global and local embeddings achieves the best overall performance. While local embeddings drive retrieval accuracy, global embeddings provide complementary, holistic context, benefiting queries that require document-wide understanding. 
This synergy further improves over local embeddings alone by 2.0 nDCG@5 points and 1.9 Recall@5 points on average, yielding the highest scores across all datasets.

\subsection{Impact of Region Rearrangement}
\label{app-layout-ablation}

\begin{table}[t]
\caption{Retrieval-augmented generation with different region rearrangement strategies. 
\ourname{} and Qwen2.5-VL-3B are used as retriever and answer generator, respectively.
}
\label{tab-layout}
\setlength{\tabcolsep}{3pt}
\resizebox{\linewidth}{!}{
\begin{tabular}{ccccccc}
\toprule
\multirow{2}{*}{\textbf{Strategy}} & \multicolumn{3}{c}{\textbf{InfoVQA}} & \multicolumn{3}{c}{\textbf{TATDQA}} \\
\cmidrule(lr){2-4} \cmidrule(lr){5-7}
                      & ANLS           & F1             & \# Token      & ANLS           & F1             & \# Token      \\
\midrule
Page-level            & 69.07          & 58.92          & 4292          & 46.10          & 31.28          & 4844          \\
Preserved Layout      & \textbf{69.48} & 58.90          & 3392          & \textbf{46.66} & 31.19          & 2502          \\
Selective Layout      & 68.28          & 57.11          & 3392          & 44.54          & 29.86          & 2502          \\
Compact Layout        & 69.34          & \textbf{59.04} & \textbf{2436} & 46.04          & \textbf{31.42} & \textbf{1839} \\
\bottomrule
\end{tabular}
}
\vskip -0.1in
\end{table}

Table~\ref{tab-layout} compares the generation performance of different region rearrangement strategies. The results show that Preserved Layout achieves the best ANLS on both datasets, slightly outperforming page-level visual augmentation.
This advantage stems from its ability to retain rich contextual relationships among critical regions while removing a portion of irrelevant visual content.
Compact Layout achieves ANLS comparable to Preserved Layout and the highest F1 on both datasets, while requiring the fewest visual tokens.
Although Selective Layout includes the same critical-region content as Compact Layout, its performance is noticeably lower. 
We attribute this degradation to the presence of extensive whitespace, which introduces visual redundancy and distracts the model's attention during generation.

\begin{table}[t]
\centering
\caption{Generation ANLS on DocVQA and DUDE by question subset. Table + numeric denotes table-related queries with numeric answers, and Other denotes the remaining queries. Region-level uses Compact Layout. $\Delta$ is Region-level minus Page-level.}
\label{tab:generation_question_subsets}
\setlength{\tabcolsep}{3pt}
\resizebox{\linewidth}{!}{
\begin{tabular}{lccc|ccc}
\toprule
\multirow{2}{*}{\textbf{Input}}
& \multicolumn{3}{c}{\textbf{DocVQA}}
& \multicolumn{3}{c}{\textbf{DUDE}} \\
\cmidrule(lr){2-4} \cmidrule(lr){5-7}
& \makecell{Table +\\numeric} & Other & Overall
& \makecell{Table +\\numeric} & Other & Overall \\
\midrule
Page-level   & 92.19 & 82.35 & 84.49 & 63.64 & 63.01 & 63.18 \\
Region-level & 87.35 & 81.44 & 82.74 & 55.37 & 61.25 & 59.66 \\
\midrule
$\Delta$     & $-4.84$ & $-0.91$ & $-1.75$ & $-8.27$ & $-1.76$ & $-3.52$ \\
\bottomrule
\end{tabular}
}
\end{table}

\begin{table*}[t]
\centering
\caption{Offline processing time (ms/page) and estimated FP32 embedding storage (KiB/page). Encoding includes post-hoc pooling for ColQwen-Pooling and pruning/merging for PRUNE-THEN-MERGE.}
\label{tab:encoding_indexing_cost}
\resizebox{\textwidth}{!}{
\begin{tabular}{lcccccccc}
\toprule
\multirow{3}{*}{\textbf{Method}}
& \multicolumn{4}{c}{\textbf{InfoVQA}}
& \multicolumn{4}{c}{\textbf{DocVQA}} \\
\cmidrule(lr){2-5} \cmidrule(lr){6-9}
& \makecell{CAR \\ (ms/page)} & \makecell{Encoding \\ (ms/page)} & \makecell{Combined \\ (ms/page)} & \makecell{Storage \\ (KiB/page)}
& \makecell{CAR \\ (ms/page)} & \makecell{Encoding \\ (ms/page)} & \makecell{Combined \\ (ms/page)} & \makecell{Storage \\ (KiB/page)} \\
\midrule
\rowcolor{gray!15}
ColQwen          & 0.00 & 0.90 & 0.90 & 361.70 & 0.00 & 0.84 & 0.84 & 375.45 \\
\midrule
ColQwen-Pooling  & 0.00 & 6.85 & 6.85 & 94.45  & 0.00 & 4.90 & 4.90 & 94.10 \\
PRUNE-THEN-MERGE & 0.00 & 6.72 & 6.72 & 91.65  & 0.00 & 4.80 & 4.80 & 95.00 \\
\ourname{}(Ours) & 1.23 & 4.68 & \textbf{5.91} & \textbf{68.80}  & 0.90 & 3.07 & \textbf{3.97} & \textbf{86.40} \\
\bottomrule
\end{tabular}
}
\end{table*}

\subsection{Generation Performance on DocVQA and DUDE}
\label{app-generation_question_subsets}

Compared with page-level augmentation, region-level augmentation yields more pronounced ANLS decreases on DocVQA and DUDE, as shown in Table~\ref{tab-generation}. Our per-query analysis identifies table-related numeric queries as the main source of the larger drops.

Table~\ref{tab:generation_question_subsets} summarizes the performance of this subset alongside the remaining queries.
For table-related numeric queries, ANLS drops by 4.84 points on DocVQA and 8.27 points on DUDE, compared with 0.91 and 1.76 points for other queries.
These questions often require row or column headers to interpret a numeric value.
Region selection and cropping can omit these headers or disrupt their association with the value, leaving the generator with insufficient context even when the value itself is retained.

\subsection{Storage and Runtime Costs}
\label{app-storage_runtime_costs}

We analyze the storage and runtime costs of \ourname{} across three stages: offline document encoding, online retrieval, and answer generation.

\paragraph{Encoding.}
Table~\ref{tab:encoding_indexing_cost} reports CAR preprocessing time, encoding time, their combined offline cost, and embedding storage on InfoVQA and DocVQA.
CAR preprocessing includes text-aware OCR and vision-based CCD.
For ColQwen-Pooling and PRUNE-THEN-MERGE, encoding time also includes their post-hoc pooling and pruning/merging operations, respectively.
We estimate embedding storage at FP32 precision as $4\bar{n}d/1024$ KiB/page, where $\bar{n}$ is the average number of document vectors and $d$ is their dimension.

\ourname{} has lower combined offline cost and smaller embedding storage than ColQwen-Pooling and PRUNE-THEN-MERGE on both datasets.
It constructs compact regions before encoding, avoiding subsequent vector pooling or pruning.
ColQwen has the lowest offline processing time among the compared methods but retains the largest document representations.

\begin{table}[t]
\caption{Comparison of average retrieval latency per query under identical environments. Latency is reported in milliseconds.}
\label{tab-retrieval-latency}
\setlength{\tabcolsep}{2pt}
\resizebox{\linewidth}{!}{
\begin{tabular}{l|ccccc}
\toprule
\multirow{1}{*}{\textbf{Method}} & \textbf{InfoVQA} & \textbf{DUDE} & \textbf{DocVQA} & \textbf{TATDQA} & \textbf{ViDoRe-P} \\
\midrule
ColQwen             & 11.38 & 61.46 & 6.29 & 0.89 & 10.53 \\
ColQwen-Pooling     & 3.17 & 25.50 & 1.84 & 0.69 & 4.15 \\
DocPruner           & 3.49 & 27.72 & 2.58 & 0.45 & 3.58 \\
Light-ColPali       & 3.32 & 23.68 & 1.55 & 0.53 & 4.13 \\
PRUNE-THEN-MERGE    & 3.51 & 24.13 & 1.64 & 0.44 & 3.89 \\
\ourname{} (Ours)   & 3.66 & 29.57 & 1.67 & 0.66 & 4.56 \\
\bottomrule
\end{tabular}
}
\end{table}

\paragraph{Retrieval.}
Table~\ref{tab-retrieval-latency} reports average online retrieval latency per query using precomputed document representations, excluding offline preprocessing and encoding.
Multi-vector retrieval uses ColPali-style exact late-interaction scoring.
\ourname{} reduces retrieval latency relative to ColQwen, while its latency ranking among efficient multi-vector methods varies across datasets.
The product of average embedding number and dimension, $\bar{n}d$, measures representation size but does not fully determine retrieval time.
Even at a fixed $\bar{n}d$, the specific dimension $d$ changes the tensor shapes and execution efficiency of similarity computation and MaxSim reduction.
Thus, fewer high-dimensional region vectors do not necessarily yield proportionally lower latency than more numerous low-dimensional patch vectors.

\begin{table}[t]
\centering
\caption{Average latency for complete answer generation (ms/query), including visual input processing. Both settings use Qwen2.5-VL-3B; region-level augmentation uses Compact Layout.}
\label{tab:generation_latency}
\setlength{\tabcolsep}{3pt}
\resizebox{\linewidth}{!}{
\begin{tabular}{lccccc}
\toprule
\textbf{Input} & \textbf{InfoVQA} & \textbf{DUDE} & \textbf{DocVQA} & \textbf{TATDQA} & \textbf{ViDoRe-P} \\
\midrule
Page-level   & 20.35 & 25.70 & 23.89 & 22.60 & 21.94 \\
\midrule
Region-level 
& \efficiencycell{12.52}{38.48\%} 
& \efficiencycell{13.87}{46.03\%} 
& \efficiencycell{10.38}{56.55\%} 
& \efficiencycell{9.45}{58.19\%}  
& \efficiencycell{10.29}{53.10\%} \\
\bottomrule
\end{tabular}
}
\end{table}

\paragraph{Generation.}
We compare page-level augmentation with region-level augmentation using Compact Layout and the same Qwen2.5-VL-3B generator.
Selecting relevant regions and compacting their layout reduces the visual input processed by the generator.
As reported in Table~\ref{tab-generation-efficiency}, the average visual token count decreases by 43.24\%--62.03\% across the five datasets.
Table~\ref{tab:generation_latency} reports average per-query latency for complete answer generation, including visual input processing, under the same hardware and decoding settings.
Region-level augmentation reduces generation latency from 20.35--25.70 ms/query to 9.45--13.87 ms/query, corresponding to reductions of 38.48\%--58.19\%.
These results show that the smaller visual input also reduces the time required to generate an answer.

\section{Qualitative Results}
\label{app-qualitative_case_studies}

\begin{figure*}
    \centering
    \includegraphics[width=\linewidth]{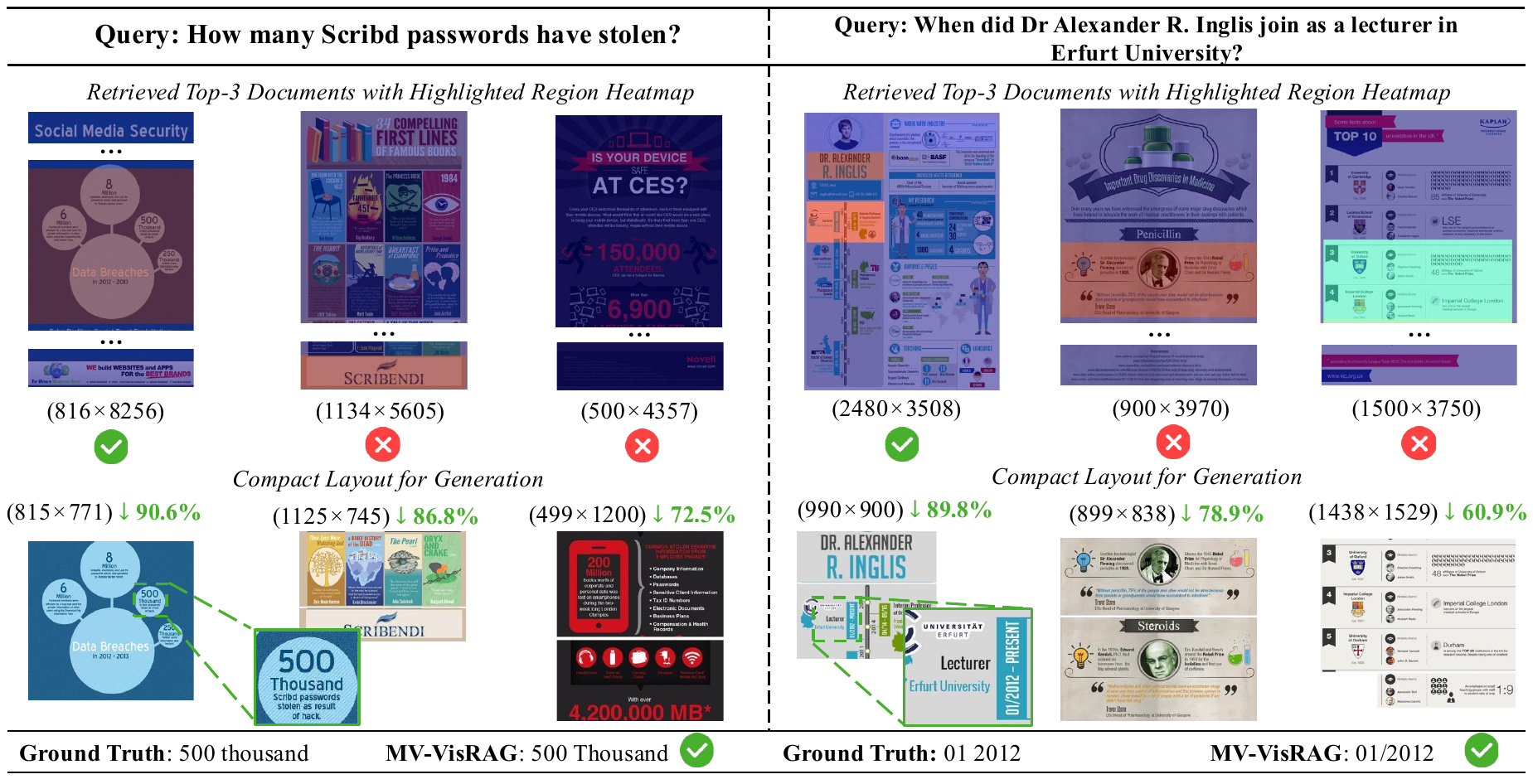}
    \caption{Strength cases from InfoVQA. For long documents, \ourname{} preserves sufficient information with a few selected regions while reducing visual context from irrelevant content.
}
    \label{fig:case_study-1}
\end{figure*}

\begin{figure*}
    \centering
    \includegraphics[width=\linewidth]{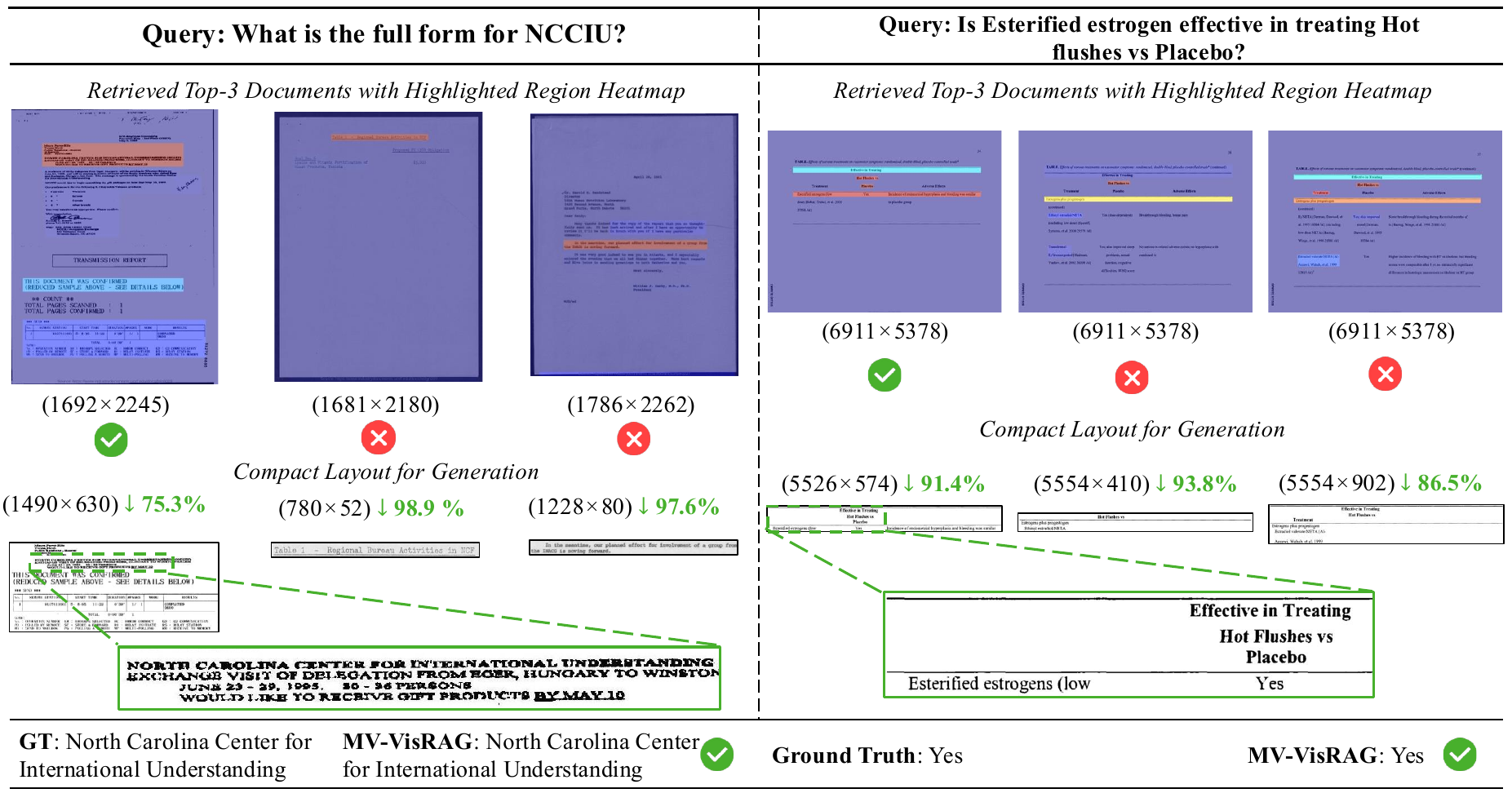}
    \caption{Strength cases from DocVQA. \ourname{} enables compact document layout, significantly reducing visual context from blank areas.}
    \label{fig:case_study-2}
\end{figure*}

\paragraph{Strength cases.}
Figures~\ref{fig:case_study-1} and~\ref{fig:case_study-2} present successful examples from InfoVQA and DocVQA. In long infographics and scanned documents, page-level visual augmentation often includes substantial irrelevant content, while the answer evidence occupies only a small portion of the page. \ourname{} identifies these regions and rearranges them into compact layouts, removing irrelevant content, blank space, and redundant visual areas. In these examples, the input image area is reduced by 60.9\%--98.9\% while preserving sufficient evidence for correct answer generation. These cases show that region-level augmentation is especially effective when the answer can be localized to a few semantically complete regions, consistent with the long-document results in Figure~\ref{fig-distribution-to-length}.

\begin{figure*}
    \centering
    \includegraphics[width=\linewidth]{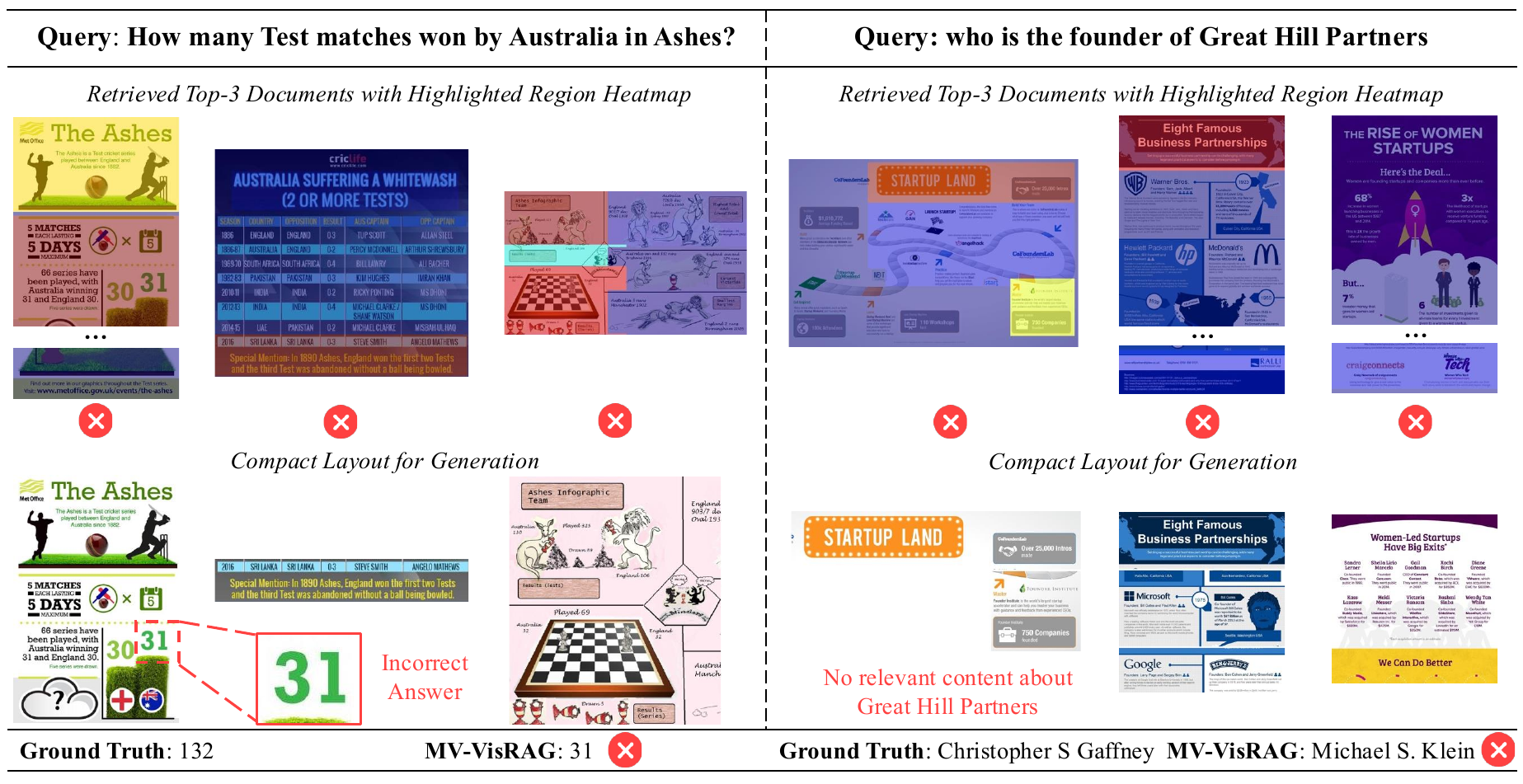}
    \caption{Retrieval failure cases. \textbf{Left:} The retriever identifies the relevant entity but not the answer-bearing content. \textbf{Right:} The retriever fails to retrieve the entity mentioned in the query.}
    \label{fig:case_study-3}
\end{figure*}

\paragraph{Retrieval failure cases.}
Figure~\ref{fig:case_study-3} presents two retrieval failure cases. 
(1) The retriever finds documents related to the Ashes, but not the answer-bearing statistic, leading to the incorrect prediction ``31'' instead of ``132''. 
(2) It fails to retrieve documents about Great Hill Partners and predicts an unrelated person. These examples show that region-level multi-vector retrieval can still confuse topical relevance with answer-bearing relevance, especially when documents share similar entities, layouts, or visual patterns.

\begin{figure*}
    \centering
    \includegraphics[width=\linewidth]{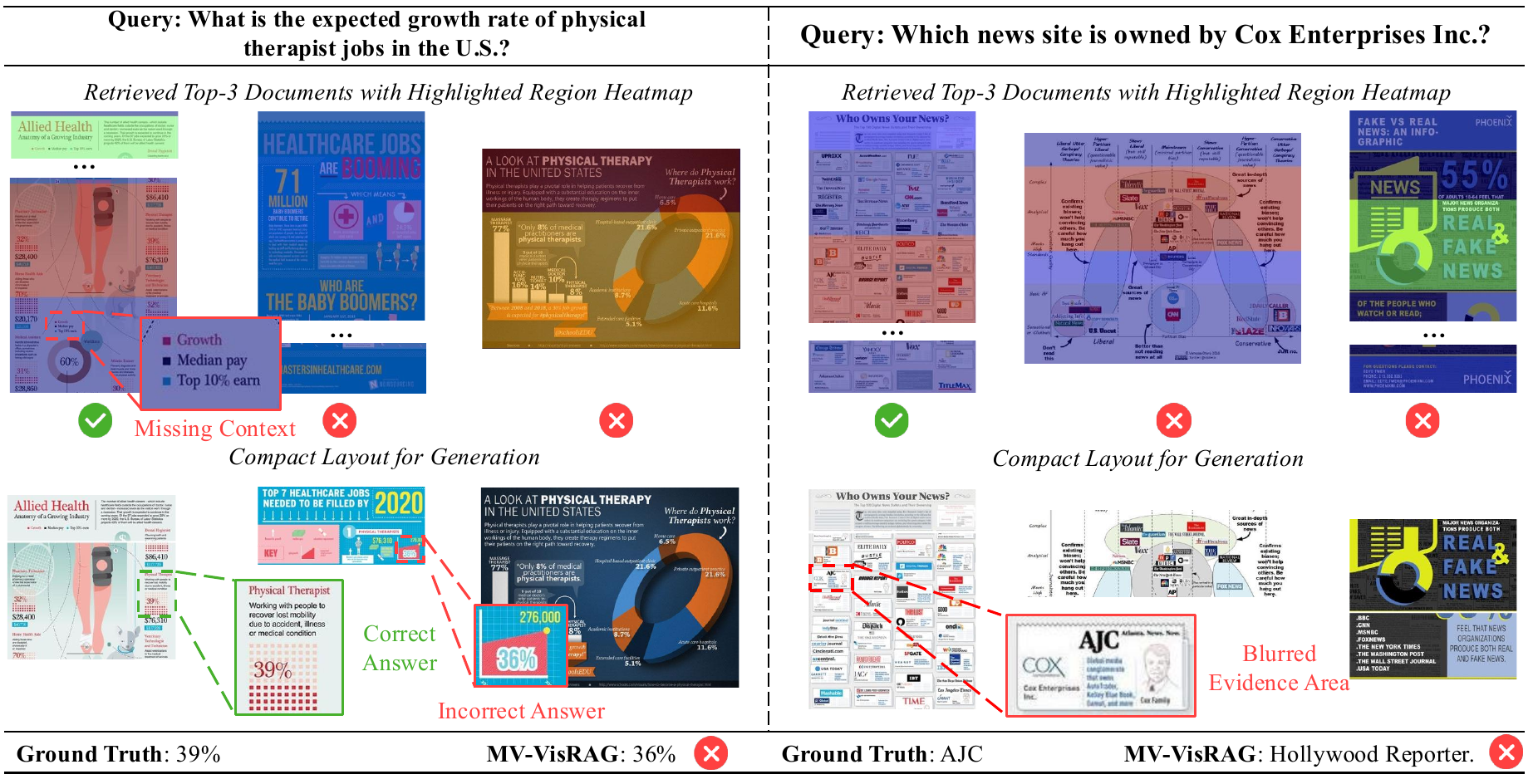}
    \caption{Generation failure cases. \textbf{Left:} The generator cannot gather sufficient information across regions because region-level augmentation misses part of the context. \textbf{Right:} The generator fails to locate the query entity or answer due to poor image quality.}
    \label{fig:case_study-4}
\end{figure*}

\paragraph{Generation failure cases.}
Figure~\ref{fig:case_study-4} shows two generation failure cases. 
(1) The evidence for the expected growth rate of physical therapist jobs is partially present, but the compact region-level context misses surrounding information needed to disambiguate nearby values, resulting in ``36\%'' instead of ``39\%''. 
(2) The answer region is relevant but blurred, causing the generator to predict ``Hollywood Reporter'' rather than ``AJC''. These cases indicate that generation may fail even with relevant regions when the compact layout omits necessary context or the visual evidence is poor.

\paragraph{Discussion.}
Overall, \ourname{} is most effective when the answer evidence is concentrated in a few semantically complete regions. The failures reflect common challenges in retrieval-based Vision-RAG systems: retrievers may confuse topical relevance with answer-bearing relevance, and generators may fail when visual evidence is incomplete, ambiguous, or lacks necessary context. These findings motivate future work on answer-evidence verification, adaptive context expansion, and evidence-quality estimation.

\end{document}